\documentclass{article}
\usepackage[main,final]{neurips_2026}
\usepackage{times}
\usepackage[utf8]{inputenc}
\usepackage[T1]{fontenc}
\usepackage{hyperref}
\usepackage{url}
\usepackage{booktabs}
\usepackage{array}
\usepackage{amsfonts}
\usepackage{nicefrac}
\usepackage{microtype}
\usepackage[table]{xcolor}  
\usepackage{amsmath}
\usepackage{amsthm}
\usepackage{graphicx}
\usepackage{subcaption}
\usepackage{multirow}
\usepackage{mathtools,bm,amssymb}
\usepackage{hyperref}
\usepackage{subcaption,placeins}
\usepackage{adjustbox,multirow,dcolumn}
\usepackage{algorithm,algpseudocode,setspace}
\usepackage{xstring}
\usepackage{cleveref}
\usepackage{graphicx}
\usepackage{natbib}
\setcitestyle{sort,comma,numbers,square}
\usepackage{tikz}
\usepackage{pgfplots}
\usepackage{comment}

\newcommand*\samethanks[1][\value{footnote}]{%
    \footnotemark[#1]\hspace{0.4em}}

\usepackage{amsmath,amsfonts,bm}

\def\1{\bm{1}}

\DeclareMathAlphabet{\mathsfit}{\encodingdefault}{\sfdefault}{m}{sl}
\SetMathAlphabet{\mathsfit}{bold}{\encodingdefault}{\sfdefault}{bx}{n}

\makeatletter
\DeclareRobustCommand\onedot{\futurelet\@let@token\@onedot}
\def\@onedot{\ifx\@let@token.\else.\null\fi}

\newcommand{\ie}{\emph{i.e\@\onedot}}

\graphicspath{{figures/}}

\usetikzlibrary{external}
\usepgfplotslibrary{colorbrewer}
\pgfplotsset{compat=newest}
\definecolor{TableBandGray}{gray}{0.94}
\newcolumntype{d}{D{.}{.}{3.2}}
\newcolumntype{C}[1]{>{\centering\arraybackslash}m{#1}}
\makeatletter
\newcolumntype{B}{>{\boldmath\DC@{.}{.}{3.2}}c<{\DC@end}}
\makeatother

\newcommand{\tabtight}{%
    \setlength{\tabcolsep}{2.6pt}%
    \renewcommand{\arraystretch}{0.92}}
\newcommand{\tabgroup}[2]{%
    \addlinespace[2pt]
    \rowcolor{TableBandGray}
    \multicolumn{#1}{@{}c@{}}{\textbf{#2}} \\
    \addlinespace[1pt]}
\newcommand{\XSolidBrunsh}{\raisebox{0.15ex}{\scalebox{0.9}{\(\boldsymbol{\times}\)}}}

\theoremstyle{definition}

\theoremstyle{plain}
\newtheorem{lemma}{Lemma}[section]

\newcommand{\hpad}{\hspace*{\fill}}
\newcommand{\subplotsizeslack}{1}
\NewDocumentCommand{\subplot}{ O{1} O{} m m m }{%
    \hpad%
    \begin{subfigure}{(\textwidth / #1) * \subplotsizeslack}
        \includegraphics[width=\textwidth, #2]{#3}
        \caption{#4}\label{#5}
    \end{subfigure}%
}

\title{{SPRM}: From Cooperative Games to Marginal-Contribution Process Reward Modeling}

\author{%
  Yu Bao\textsuperscript{1,2}\thanks{Equal contribution.}
  \And
  Pak Lon Ip\textsuperscript{1}\samethanks[1]
  \And
  Qiyu Ruan\textsuperscript{1,2}
  \And
  Xitong Gao\textsuperscript{3}\thanks{Corresponding authors:
    \texttt{xt.gao@siat.ac.cn} and \texttt{czxu@um.edu.mo}.}
  \And
  Chengzhong Xu\textsuperscript{1,2}\samethanks[2]
  \AND
  \normalfont
  \textsuperscript{1}~SKL of Internet of Things for Smart City,
  University of Macau.\\
  \textsuperscript{2}~Department of Computer Science, University of Macau.\\
  \textsuperscript{3}~Shenzhen Institutes of Advanced Technology, Chinese Academy of Sciences.
}

\begin{document}

\maketitle

\begin{abstract}

Process reward models provide step-level feedback for reasoning,
but reliable process supervision remains difficult to obtain:
human annotations are costly, Monte Carlo rollouts require large sampling budgets,
and LLM judges can introduce prompt-sensitive biases.
We introduce SPRM, an outcome-only framework that
constructs process rewards directly from terminal correctness signals.
SPRM views a reasoning trajectory as a temporally ordered cooperative game,
where reasoning steps act as participants and the final outcome reward is the shared payoff.
Following this allocation view, SPRM learns a differentiable trajectory value function
from outcomes and distributes the baseline-relative payoff difference to steps
through Aumann-Shapley-style Step Value Integration.
We further analyze its Shapley properties under the temporal and causal structure
of reasoning trajectories.
To stabilize supervision, SPRM incorporates model-counterfactual consistency
and estimates a cross-trajectory inherent score that captures each step's
context-independent value from semantically similar steps.
These signals train a dual-head process reward model that jointly represents
trajectory-specific marginal credit and cross-trajectory inherent step value.
Across ProcessBench, PRMBench, and Best-of-N reranking on AceMath-RewardBench,
SPRM improves first-error localization, marginal-contribution-sensitive evaluation,
and test-time reasoning selection over existing open PRMs trained from human,
rollout, curated, or outcome-broadcast supervision.
These results suggest that marginal-contribution-based credit redistribution
offers an effective and scalable route to process supervision from outcome-only data.
Our code is available at
\url{https://github.com/by0214/sprm-open}.
\end{abstract}

\section{Introduction}
Large language models (LLMs) have demonstrated strong reasoning ability on 
tasks such as mathematical proof generation \citep{DBLP:math1,DBLP:math2,DBLP:math3,DBLP:math4}, 
interactive decision making \citep{DBLP:decision1}, and
code synthesis \citep{DBLP:code1,DBLP:code2}.
Yet their training and alignment are still dominated by outcome-level signals:
rewards are typically assigned only to the final answer, while
intermediate steps in reasoning chains receive little direct credit.
This sparse-reward setting creates a severe temporal credit-assignment
problem \citep{DBLP:journals/tmlr/PignatelliFGMHT24,DBLP:credit1,DBLP:credit2}.
A model can observe whether a trajectory eventually succeeds, but it cannot
directly tell which steps contributed to that success.
Classical reinforcement learning methods such as PPO \citep{DBLP:PPO} and reasoning-oriented
variants such as GRPO \citep{DBLP:GRPO} therefore still optimize mainly from outcome-level
rewards, while reward-free preference optimization methods such as DPO \citep{DBLP:DPO} learn
from trajectory-level preferences over final outcomes.
Neither paradigm by itself identifies which intermediate reasoning steps
deserve credit.
As a result, process reward models (PRMs) \citep{xi2025agentprmprocessrewardmodels,lin2025toolprmfinegrainedinferencescaling,khalifa2025think}, 
which learn step-level reward signals for reasoning processes, have become an important component in
test-time scaling, reranking, and reinforcement learning for LLM reasoning.

\begin{figure*}[t]
\centering
\includegraphics[width=\textwidth]{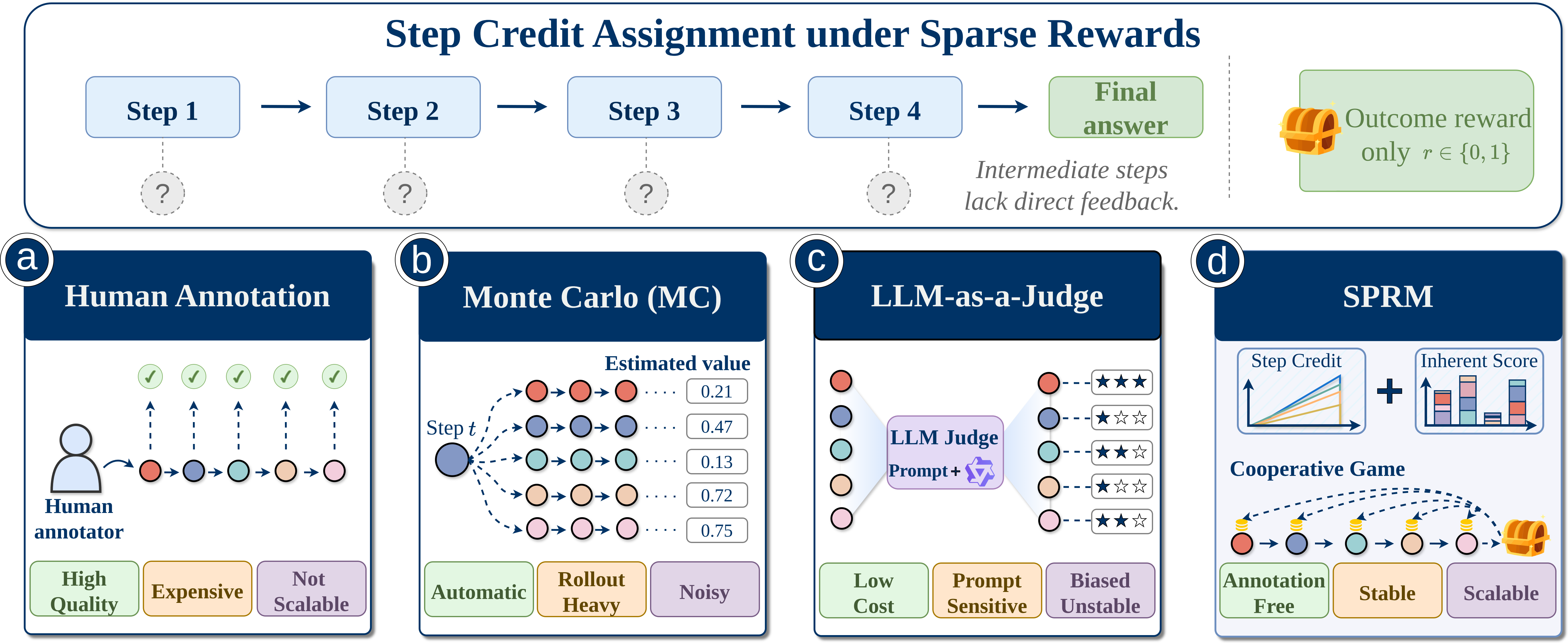}
\caption{Comparison of PRM supervision under sparse outcome rewards.
(a) Human annotation is reliable but costly, while (b) rollout-based
estimation and (c) LLM judging provide noisy and unstable supervision.
Unlike these external estimators, SPRM derives step-level supervision from
outcomes through a cooperative-game view.}
\label{fig:supervision_compare}
\end{figure*}

The main challenge in developing current process reward models is not the
model architecture, but obtaining reliable training data for assigning credit
to individual reasoning steps.
Human annotation \citep{uesato2022solving, lightman2023lets} is reliable but expensive and difficult to scale.
Monte Carlo-based labeling \citep{wang2024math,zhang2025process} is more automatic and therefore less expensive,
but it requires large rollout budgets and often remains noisy, since the
same step can receive different labels across sampled continuations and
value estimates.
LLM-as-a-judge \cite{zeng2025versaprm, duan2025efficient,yang2025beyond,wang2025athena,cao2025dreamprm} 
further reduces annotation cost, yet its labels can inherit
biases from pretraining data and be sensitive to prompt wording or scoring
criteria, which makes the resulting credits unstable.
As illustrated in \Cref{fig:supervision_compare}, these approaches share a
common mismatch: they import external annotation, rollout, or judge
distributions to explain trajectories whose only observed training signal is
the final outcome.
This dependence limits robustness, scalability, and reusability, and motivates
the question of whether reliable step credit can be derived directly from
trajectory data itself.

Beyond the dependence on external supervision, we argue that a good
step-level credit assignment should also account for each step's marginal
contribution to the final trajectory reward.
In a successful reasoning chain, a decisive transformation, a harmless
restatement, and an irrelevant detour are all paired with the same terminal
reward, but they should clearly receive different credit.
A contribution-faithful signal should therefore concentrate the final reward on
the steps that actually support the outcome, assign little credit to steps whose
presence does not change the result, and avoid changing its judgment merely
because reasoning steps appear in different positions.
We therefore formulate the reasoning process as a cooperative game in which
steps participate in the allocation while retaining their temporal dependencies:
each step is a participant, the
terminal reward is the shared payoff, and credit should be determined by
contribution rather than temporal position.
From this perspective, sparse outcome supervision becomes a question of payoff
allocation: how should a terminal reward be redistributed into principled
step-level credit?

Shapley values provide a principled starting point because they allocate payoff
by marginal contribution, making them well matched to the goal of converting a
sparse final reward into step-level credit.
However, reasoning trajectories differ from the classical Shapley setting in a
crucial way: steps are temporally ordered, causally dependent, and not freely
exchangeable.
Under these constraints, a naive Shapley construction can degenerate into
prefix-wise value differences along the original sequence, making the resulting
credit highly sensitive to the context in which a step appears and often still
dependent on strong external value estimators or evaluators.
The central challenge is therefore to obtain Shapley-style, contribution-faithful
credit from terminal outcome signals alone while respecting the temporal-causal
structure of reasoning.

Motivated by this goal, we propose \textbf{SPRM}
(\textbf{S}hapley-inspired \textbf{P}rocess \textbf{R}eward
\textbf{M}odel), a framework that automatically constructs PRM training
signals from outcome-only rewards.
SPRM uses the idea of Aumann-Shapley, a continuous-space variant of Shapley
values, to perform marginal-contribution-based credit allocation over reasoning
steps.
Combined with model-counterfactual consistency, this allocation captures how a
step changes the value of its own reasoning trajectory while accounting for
interactions among reasoning steps.
It further uses a cross-trajectory inherent score to improve the stability of
PRM supervision for steps across different contexts.

SPRM directly addresses a central limitation of existing PRM data-construction
pipelines: obtaining reliable step-level supervision without costly human
annotation or strong external evaluators.
SPRM instead extracts PRM supervision directly from sparse terminal rewards,
thereby removing dependence on strong external estimators.
This provides an automated and scalable route to process supervision while
preserving the temporal grounding needed for multi-step reasoning.

Our main contributions are summarized as follows:

\begin{itemize}
    \item We introduce an outcome-only PRM data-construction framework that
    derives step-level training labels from terminal correctness rewards,
    without human step labels, rollout labels, LLM judges, or pretrained
    process evaluators.
    \item We instantiate marginal-contribution supervision through Step Value
    Integration, which attributes a learned differentiable trajectory value to
    individual reasoning steps under a temporally constrained cooperative-game
    view.
    \item We add two stabilizing supervision signals, model-counterfactual
    consistency and cross-trajectory inherent scoring, and validate the resulting
    supervision through controlled same-backbone comparisons, process evaluation,
    Best-of-N reranking, downstream reinforcement learning, and experiments in
    verifiable code generation.
\end{itemize}

\section{Related Work}
\textbf{Process Reward Model}
Process reward models (PRMs) provide fine-grained reward signals at each
reasoning step and have substantially improved LLM performance on complex
multi-step tasks such as mathematical problem solving and code generation
\citep{uesato2022solving, lightman2023lets}.
Compared with outcome reward models that supervise only the final answer, PRMs
can localize intermediate errors more precisely, improve temporal credit
assignment, and mitigate the alignment gap where the answer is correct but the
reasoning process is flawed.
As a result, PRMs have become a core technique for test-time scaling, such as
Best-of-N \citep{lin2025toolprmfinegrainedinferencescaling}, and
reinforcement learning
\citep{DBLP:conf/iclr/LiL25a, xi2025agentprmprocessrewardmodels, mit2025process}.
Most PRMs are trained to judge whether an intermediate step is locally correct,
but correctness alone does not specify how much credit the step deserves for
the final outcome.
SPRM complements this view by introducing a fairness-oriented credit criterion:
it measures step credit through marginal contribution, so that process rewards
reflect how each reasoning step changes the trajectory outcome rather than only
whether the step appears plausible in isolation.

\textbf{Automated Annotation and Credit Redistribution}
Automated process-labeling methods include MC-based step annotation
\citep{wang2024math, zhang2025process, zhang-etal-2025-lessons, chen-etal-2025-better-process, ding2025scan}
and LLM-as-a-judge labeling \citep{zeng2025versaprm, duan2025efficient}.
MC labels require large rollout budgets and can be noisy, while judge labels
can inherit model biases and prompt sensitivity.
Hybrid pipelines combine the two to balance quality and scalability
\citep{wang2025athena, zhang-etal-2025-lessons}, but still require extra
rollouts or judge calls.
Credit-assignment methods for delayed rewards provide a complementary view:
RUDDER \citep{arjona2019rudder} redistributes terminal returns through
contribution analysis while preserving the trajectory objective in a
return-equivalent decision process.
Shapley-based model explanations offer a principled allocation rule, but their
outputs depend on how the value function, background distribution, and
explanation context are operationalized \citep{pmlr-v119-sundararajan20b}.
SPRM instead learns an internal differentiable value function and assigns
Aumann-Shapley-style credit from outcome-only trajectories, avoiding external
supervision or attribution sources such as rollouts, LLM judges, pretrained
evaluators, or background explanation distributions.

\section{Learning SPRM from Outcome-Only Trajectories}
\subsection{Problem Formulation and Motivation}
\label{sec:formulation}

An outcome-supervised reasoning example consists of a question or instruction
\(q_i\), a sequence of intermediate reasoning steps
\(s_{i,1},\ldots,s_{i,T_i}\), and a terminal ground-truth reward
\(r_i\in\{0,1\}\).
We write the resulting trajectory as
\(
\tau_i=(q_i,s_{i,1},\ldots,s_{i,T_i})
\)
and the dataset as
\(
\mathcal{D}=\{(\tau_i,r_i)\}_{i=1}^{N}
\).

In an ordered reasoning trajectory, a step's apparent utility is entangled with
its position: later steps can benefit from a correct prefix, while earlier steps
may be under-credited once later steps complete the solution.
We therefore treat outcome-only supervision as a fair credit-allocation problem,
where the trajectory reward is produced by cooperation among steps and should be
assigned by marginal contribution rather than temporal artifacts.

Following \citet{shapley1953value}, we define a finite cooperative game as a
pair \((N,v)\), where \(N\) is the participant set and
\(v:2^N\rightarrow\mathbb{R}\) gives the payoff \(v(S)\) of each coalition
\(S\subseteq N\), with \(v(\emptyset)=0\).
For a fixed participant set \(N\), an allocation rule \(\psi\) maps a value function
\(v\) to a credit vector \(\psi(v)=(\phi_j(v))_{j\in N}\), where
\(\phi_j(v)\) is the credit assigned to participant \(j\).
We call an allocation fair if it satisfies the four Shapley axioms in
\Cref{tab:classical_shapley_axioms}.

\begin{table}[h]
\centering
\caption{Classical Shapley axioms for a finite cooperative game.}
\label{tab:classical_shapley_axioms}
\small
\begin{tabular}{p{0.21\linewidth}p{0.69\linewidth}}
\toprule
Axiom & Meaning \\
\midrule
Efficiency &
The total payoff is fully allocated:
\(\sum_{j\in N}\phi_j(v)=v(N)-v (\emptyset)\). \\
Symmetry &
If two participants make the same marginal contribution to every coalition, they
receive the same credit. \\
Dummy participant &
If a participant never changes any coalition value, its credit is zero. \\
Additivity / Linearity &
For two games \(v\) and \(w\), the allocation for \(av+bw\) is
\(a\phi(v)+b\phi(w)\). \\
\bottomrule
\end{tabular}
\end{table}

To connect this game-theoretic formulation to reasoning, we map the reasoning
steps in trajectory \(\tau_i\) to the participant set
\(
\mathcal{N}_i=\{s_{i,1},\ldots,s_{i,T_i}\}
\),
and use the trajectory outcome as the shared payoff.
For a coalition \(S\subseteq\mathcal{N}_i\) with value \(v(S)\), the classical
Shapley credit of step \(s_{i,t}\) is
\begin{equation}
\label{eq:discrete_shapley}
\phi_{i,t}^{\mathrm{sha}}
:=
\frac{1}{T_i!}
\sum_{\pi\in\Pi(\mathcal{N}_i)}
\Big[
v\bigl(\mathrm{Pre}_{\pi}(s_{i,t})\cup\{s_{i,t}\}\bigr)
-
v\bigl(\mathrm{Pre}_{\pi}(s_{i,t})\bigr)
\Big],
\end{equation}
where \(\Pi(\mathcal{N}_i)\) is the set of permutations and
\(\mathrm{Pre}_{\pi}(s_{i,t})\) denotes predecessors of \(s_{i,t}\) under
\(\pi\).

Shapley allocation is appealing because, in unordered cooperative games, it
provides a fair marginal-contribution allocation.
Reasoning trajectories, however, are not exchangeable coalitions: their steps
must respect temporal-causal dependencies, so the classical symmetry axiom
cannot be imposed literally.
We call this restricted setting a \emph{temporally constrained reasoning game},
which assumes the temporal-order condition in
\Cref{sec:appendix_assumptions}, and define its
\emph{admissible-order Shapley-style credit} as the Shapley
marginal-contribution average taken only over temporally admissible orders; see
\Cref{sec:appendix_shapley_axioms} for the resulting gap from classical Shapley
symmetry.
Let
\(
P_{i,t}=\{s_{i,1},\ldots,s_{i,t}\}
\)
and \(P_{i,0}=\emptyset\).
\begin{lemma}
\label{lem:temporal_shapley_prefix}
In a temporally constrained reasoning game where the only admissible order is
\(1\prec2\prec\cdots\prec T_i\), the admissible-order Shapley-style credit of
step \(s_{i,t}\) is the prefix value difference
\begin{equation}
\label{eq:value_difference_reward}
\phi_{i,t}^{\mathrm{ord}}
=
v(P_{i,t})-v(P_{i,t-1}).
\end{equation}
\end{lemma}
\Cref{sec:appendix_step_credit_proofs} provides the proof.
The lemma shows that this ordered construction no longer averages over multiple
orders, but collapses to prefix differences.
This form is operationally unattractive: it requires accurate multi-step value
regression over prefixes, and any prediction error can propagate along the
reasoning chain.
We therefore move from discrete prefix differences to a continuous allocation
view based on Aumann-Shapley values \citep{aumann1974values}, which generalize
Shapley's average over permutations to the continuous setting for measuring
marginal contribution.
Integrated Gradients provides a concrete instance of this idea and is the
unique continuous attribution rule satisfying Shapley-aligned axioms
\citep{sundararajan2017axiomatic,pmlr-v119-sundararajan20b}.
Building on this principle, we propose \textbf{Step Value Integration} (SVI): a lightweight
outcome-trained trajectory value function whose gradients are integrated to
assign step credit.

\subsection{Step Value Integration with Model-Counterfactual Consistency}
\label{sec:method_step}
\begin{figure*}[t]
    \centering
    \includegraphics[width=\textwidth]{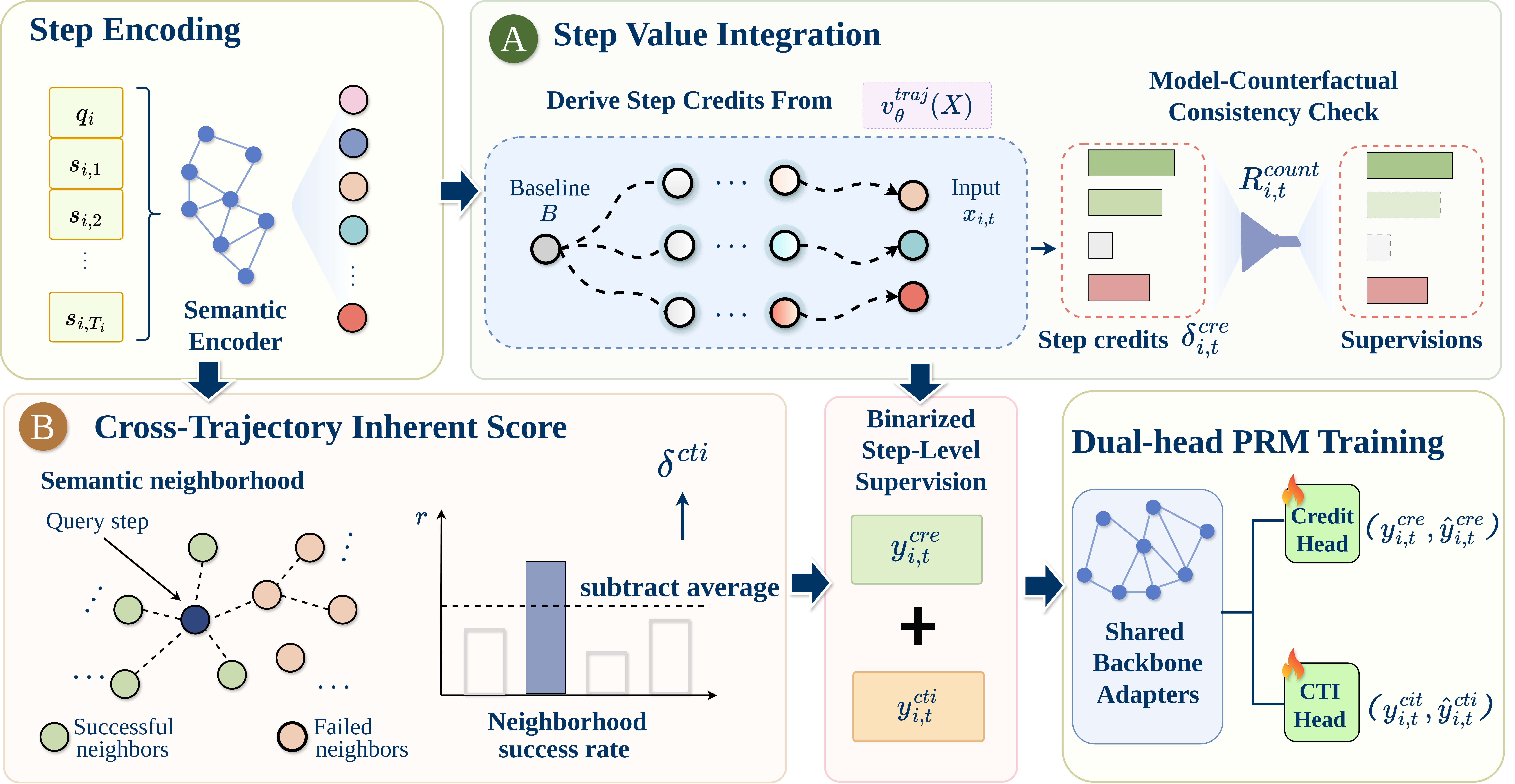}
    \caption{Overview of SPRM.
    From outcome-only trajectories, SPRM encodes questions and steps, derives
    complementary SVI and CTI supervision signals, binarizes them to train a
    dual-head PRM, and combines both heads at inference time for step rewards.}
    \label{fig:sprm_overview}
    \end{figure*}

For each trajectory
\(\tau_i=(q_i,s_{i,1},\ldots,s_{i,T_i})\), we use an encoder with sufficient
semantic encoding capacity, such as an embedding model or an LLM, to encode the
question and steps as
\(
X_i=\mathrm{encoder}(\tau_i)=[x_{i,0},x_{i,1},\ldots,x_{i,T_i}]
\in\mathbb{R}^{(T_i+1)\times d}
\),
where \(x_{i,0}\) is the question representation and \(x_{i,t}\) represents
step \(s_{i,t}\), so \(X_i\) can be viewed as the participants
\(\mathcal{N}_i\) in this game.
Note that we also encode the question itself; we refer to its representation
\(x_{i,0}\) as \textit{question\_end} and discuss its role in
\Cref{tab:ablation_processbench}.
We train a lightweight differentiable trajectory value function
\(v^{\mathrm{traj}}_{\theta}:\mathbb{R}^{(T_i+1)\times d}\rightarrow[0,1]\)
from final rewards only:
\begin{equation}
\label{eq:stage1_objective}
\theta^\star
:=
\arg\min_{\theta}
\mathbb{E}_{(\tau,r)\sim\mathcal{D}}
\left[
\mathrm{BCE}\!\left(
r,
v^{\mathrm{traj}}_{\theta}(X)
\right)
\right].
\end{equation}

The trajectory value function converts step-by-step value estimation into a
global value prediction problem, making the target learnable by a lightweight
model and removing the need for a strong external value model.
SVI assigns each step a share of the change in predicted trajectory value from
the baseline to the observed trajectory.
Let \(b\in\mathbb{R}^{d}\) be a semantically empty baseline, let
\(B_i=[b,\ldots,b]\), and define the straight-line path function
\(\gamma_i(\alpha)=B_i+\alpha(X_i-B_i)\).
The Aumann-Shapley-style credit of step \(s_{i,t}\) is
\begin{equation}
\label{eq:delta_step}
\delta_{i,t}^{\mathrm{cre}}
:=
\left(x_{i,t}-b\right)^\top
\int_0^1
\nabla_{x_{i,t}}
v^{\mathrm{traj}}_{\theta^\star}\!\left(\gamma_i(\alpha)\right)
\,d\alpha .
\end{equation}
The straight-line path specifies how credit is allocated; unlike a causal
intervention, it does not require intermediate states to correspond to valid
textual reasoning.
This attribution is interpreted under the assumptions in
\Cref{sec:appendix_assumptions}.
We analyze the advantages of this SVI credit over the naive Shapley
delta-value form in \Cref{tab:ablation_processbench}.
\begin{lemma}
\label{lem:svi_shapley_like_properties}
For a fixed trajectory value function and baseline, with the straight-line path
function \(\gamma_i\), the SVI credits in \Cref{eq:delta_step} satisfy
Shapley-like properties.
\end{lemma}
The proof is provided in \Cref{sec:appendix_step_credit_proofs}, and the
Shapley-like properties are discussed in \Cref{sec:appendix_shapley_axioms}.
In nonlinear value models, SVI distributes interaction credit across steps and
can be affected by local sensitivity and numerical integration noise.
We therefore introduce model-counterfactual consistency, a
representation-ablation check within the learned trajectory value model.
It combines the direct removal response, which reflects interactions involving
the current step, with pairwise interaction evidence from preceding steps in a
local window to filter SVI labels by sign agreement.
Let \(X_i^{(-t)}\) replace \(x_{i,t}\) with \(b\), and let
\(X_i^{(-t,-u)}\) replace both \(x_{i,t}\) and \(x_{i,u}\).
\begin{equation}
\label{eq:counterfactual_effects}
\begin{aligned}
R_{i,t}^{\mathrm{dir}}
&= v^{\mathrm{traj}}_{\theta^\star}(X_i)
- v^{\mathrm{traj}}_{\theta^\star}(X_i^{(-t)}),
\quad
R_{i,(t,u)}^{\mathrm{int}}
= R_{i,t}^{\mathrm{dir}} 
- v^{\mathrm{traj}}_{\theta^\star}(X_i^{(-u)})
+ v^{\mathrm{traj}}_{\theta^\star}(X_i^{(-t,-u)}).
\end{aligned}
\end{equation}
\(R_{i,t}^{\mathrm{dir}}\) measures the predicted value difference before and
after ablating step \(t\)'s representation, whereas
\(R_{i,(t,u)}^{\mathrm{int}}\) measures the pairwise interaction between step
representations \(t\) and \(u\).
We define a local window
\(\mathcal{W}(t)=\{u \mid 1 \le u < t,\; t-u \le w\}\)
to restrict the view of each step to nearby previous steps, and combine the
interaction terms with \(R_{i,t}^{\mathrm{dir}}\) to obtain the final
consistency score:
\begin{equation}
\label{eq:pairwise_counterfactual}
R_{i,t}^{\mathrm{count}}
:=
R_{i,t}^{\mathrm{dir}}
+
\beta
\frac{
\sum_{u\in\mathcal{W}(t)}
\mathbf{1}\!\left(R_{i,(t,u)}^{\mathrm{int}}>0\right)\,
R_{i,(t,u)}^{\mathrm{int}}
}{
\sum_{u\in\mathcal{W}(t)}
\mathbf{1}\!\left(R_{i,(t,u)}^{\mathrm{int}}>0\right)
+ \epsilon
}.
\end{equation}
We omit negative pairwise interactions because they may repeatedly penalize
later steps for the same upstream error, even when those steps are locally
coherent.
The direct term \(R_{i,t}^{\mathrm{dir}}\) still retains negative evidence from
representation ablation.
As a conservative consistency filter, we retain an SVI label only when
\(\delta_{i,t}^{\mathrm{cre}}\) and \(R_{i,t}^{\mathrm{count}}\) share the same
nonzero sign, allowing both positive and negative labels.

\subsection{Cross-Trajectory Inherent (CTI) Score}
\label{sec:method_cti}

In temporal-causal reasoning, each step is interpreted under its current
context, so its credit can be strongly affected by the surrounding trajectory.
For example, a useful step may be underestimated when it appears after an
incorrect prefix.
To reduce this variance, CTI estimates the inherent latent value of a step by
comparing semantically similar steps across trajectories and measuring their
association with final success.
We use this inherent score as auxiliary supervision for more stable PRM
training.
For step representation \(x_{i,t}\), retrieve top-\(M\) neighbors from other
trajectories and denote them by \(\mathcal{H}_M(i,t)\).
With similarity weights \(\omega_{i,t,j,\ell}\ge 0\) and
\(\epsilon_\omega>0\), define
\begin{equation}
\label{eq:neighborhood_success_rate}
\hat r_{i,t}
:=
\frac{
\sum_{(j,\ell)\in\mathcal{H}_M(i,t)} \omega_{i,t,j,\ell} r_j
}{
\sum_{(j,\ell)\in\mathcal{H}_M(i,t)} \omega_{i,t,j,\ell}
+\epsilon_\omega
}.
\end{equation}
Comparing this neighborhood success rate with the dataset average
\(
\bar r=\frac{1}{N}\sum_{i=1}^{N} r_i
\)
gives the cross-trajectory inherent score
\begin{equation}
\label{eq:delta_cti}
\delta_{i,t}^{\mathrm{cti}}
:=
\hat r_{i,t}-\bar r .
\end{equation}
Positive \(\delta_{i,t}^{\mathrm{cti}}\) means that the step pattern appears in
more successful neighborhoods than average.
This cross-trajectory inherent score reshapes the final step reward away from
noisy single-trajectory evidence.

\subsection{PRM Training and Inference}
\label{sec:method_prm}

The PRM uses a shared backbone with step-credit and cross-trajectory heads,
parameterized by \(\theta_{\mathrm{PRM}}\).
For step \(s_{i,t}\), the heads predict
\(\hat y_{i,t}^{\mathrm{cre}}\in(0,1)\) and
\(\hat y_{i,t}^{\mathrm{cti}}\in(0,1)\).
We binarize the constructed signals as
\(y_{i,t}^{\mathrm{cre}}=\mathbf{1}\!\left(\delta_{i,t}^{\mathrm{cre}}>0\right)\)
and
\(y_{i,t}^{\mathrm{cti}}=\mathbf{1}\!\left(\delta_{i,t}^{\mathrm{cti}}>0\right)\).
The step-credit label is used only when the attribution sign agrees with the
model-counterfactual consistency sign:
\begin{equation}
\label{eq:step_credit_mask}
m_{i,t}
:=
\mathbf{1}\!\left(
\delta_{i,t}^{\mathrm{cre}} \cdot R_{i,t}^{\mathrm{count}} > 0
\right).
\end{equation}
The final objective is
\begin{equation}
\label{eq:stage3_objective}
\theta_{\mathrm{PRM}}^\star
:=
\arg\min_{\theta_{\mathrm{PRM}}}
\mathbb{E}_{(\tau_i,r_i)\sim\mathcal{D}}
\left[
\sum_{t=1}^{T_i}
m_{i,t}\,\mathrm{BCE}\!\left(y_{i,t}^{\mathrm{cre}}, \hat y_{i,t}^{\mathrm{cre}}\right)
+
\sum_{t=1}^{T_i}
\mathrm{BCE}\!\left(y_{i,t}^{\mathrm{cti}}, \hat y_{i,t}^{\mathrm{cti}}\right)
\right].
\end{equation}
At inference time, the final step reward is
\begin{equation}
\label{eq:final_step_reward}
\hat y_{i,t}
:=
\lambda \hat y_{i,t}^{\mathrm{cre}}
+
(1-\lambda)\hat y_{i,t}^{\mathrm{cti}},
\end{equation}
where \(\lambda\in[0,1]\) controls the trade-off between trajectory-internal
step credit and cross-trajectory inherent scoring.

\section{Experiments}
\subsection{Experimental Settings}
\paragraph{Evaluation Datasets and Metrics}
Our experimental design evaluates SPRM along three complementary dimensions:
\textbf{first-error localization}, which tests whether a PRM can locate the
first wrong step in an erroneous chain while recognizing all-correct chains;
\textbf{credit assignment}, which probes whether reward reflects marginal
contribution rather than redundant or logically flawed steps; and
\textbf{reasoning-time selection}, which tests whether process rewards select
better complete reasoning trajectories during Best-of-N inference
\citep{lightman2023lets}.
We instantiate these dimensions with ProcessBench
\citep{zheng2025processbench}, PRMBench \citep{DBLP:prmbench}, and Best-of-N
reranking, respectively.
For the latter, we rerank trajectories from AceMath-RewardBench
\citep{DBLP:AceMath} and report final-answer accuracy across problem
difficulties and search ranges, averaged over 10 random seeds.
We visualize four representative datasets in the main text, aggregate step scores
according to each model's originally recommended configuration, and
defer metric definitions and complete dataset-wise results to
\Cref{sec:appendix_eval_metrics,sec:appendix_bon_results}.

\paragraph{Baselines}
We select representative baselines by supervision source, as summarized in
\Cref{tab:baseline_supervision_compare}: human-labeled PRMs
(Qwen2.5-Math-7B-PRM800K~\citep{zhang-etal-2025-lessons} and
ReasonEval-7B~\citep{DBLP:ReasonEval}), Monte Carlo-labeled PRMs
(Math-Shepherd-7B~\citep{wang2024math} and
RLHFlow~\citep{dong2024rlhf}), outcome-reward broadcasting
(OVM-Qwen3-4B~\citep{ovm:conf/naacl/YuGW24}), first-error boundary modeling
(PQM-DeepSeek-zeta4-7B~\citep{DBLP:conf/iclr/LiL25a}), and curated process
supervision (Skywork-PRM-7B~\citep{skywork}).
SPRM is trained from Math-Shepherd trajectories \citep{wang2024math} without
using its intermediate step rewards.

\subsection{Main Results}
\begin{table*}[t!]
\centering
\footnotesize
\tabtight
\caption{ProcessBench results.
\(\Delta = \mathrm{SPRM} - \mathrm{baseline}\) for the average F1;
positive values indicate higher scores for SPRM.
Data for models marked with an asterisk (\(*\)) are sourced from
\citep{zhang-etal-2025-lessons}.}
\label{tab:processbench_res}
\resizebox{0.97\textwidth}{!}{%
\begin{tabular}{@{}l*{12}{C{2.75em}}C{4.0em}C{2.9em}@{}}
\toprule
\multirow{2}{*}{\textbf{Model}} & \multicolumn{3}{c}{\textbf{GSM8K}} & \multicolumn{3}{c}{\textbf{MATH}} & \multicolumn{3}{c}{\textbf{OLYMPIADBENCH}} & \multicolumn{3}{c}{\textbf{OMNIMATH}} & \multirow{2}{*}{\textbf{Avg\_F1}} & \multirow{2}{*}{\(\mathbf{\Delta}\)} \\
\cmidrule(lr){2-4}\cmidrule(lr){5-7}\cmidrule(lr){8-10}\cmidrule(lr){11-13}
 & Err & Cor & \textbf{F1} & Err & Cor & \textbf{F1} & Err & Cor & \textbf{F1} & Err & Cor & \textbf{F1} & & \\
\midrule
\tabgroup{15}{Closed-source judges}
GPT-4o* & 70.0 & 91.2 & 79.2 & 54.4 & 76.6 & 63.6 & 45.8 & 58.4 & 51.4 & 45.2 & 65.6 & 53.5 & 61.9 & -1.9 \\
o1-mini* & 88.9 & 97.9 & 93.2 & 83.5 & 95.1 & 88.9 & 80.2 & 95.6 & 87.2 & 74.8 & 91.7 & 82.4 & 87.9 & -27.9 \\
\midrule
\tabgroup{15}{Open-source judges}
Llama-3.3-70B-Instruct* & 72.5 & 96.9 & 82.9 & 43.3 & 83.2 & 59.4 & 31.0 & 94.1 & 46.7 & 28.2 & 90.5 & 43.0 & 58.0 & \textbf{+2.0} \\
Qwen2.5-Math-72B-Instruct* & 49.8 & 96.9 & 65.8 & 36.0 & 94.3 & 52.1 & 19.5 & 97.3 & 32.5 & 19.0 & 96.3 & 31.7 & 45.5 & \textbf{+14.5} \\
Qwen2.5-72B-Instruct* & 62.8 & 96.9 & 76.2 & 46.3 & 93.1 & 61.8 & 38.7 & 92.6 & 54.6 & 36.6 & 90.9 & 52.2 & 61.2 & -1.2 \\
\midrule
\tabgroup{15}{PRMs}
Math-Shepherd-7B & 33.1 & 88.7 & 48.2 & 17.8 & 84.1 & 29.4 & 15.0 & 68.2 & 24.6 & 15.2 & 74.2 & 25.2 & 31.9 & \textbf{+28.1} \\
RLHFlow-Mistral-8B & 34.2 & 98.4 & 50.9 & 21.9 & 72.4 & 33.6 & 8.6 & 47.8 & 14.6 & 9.6 & 45.6 & 15.9 & 28.8 & \textbf{+31.2} \\
RLHFlow-DeepSeek-8B & 24.6 & 98.4 & 39.4 & 22.6 & 78.1 & 35.0 & 10.3 & 51.3 & 17.1 & 10.3 & 51.9 & 17.2 & 27.2 & \textbf{+32.8} \\
Skywork-PRM-7B & 58.4 & 70.0 & 63.7 & 40.7 & 62.3 & 49.3 & 16.0 & 20.0 & 17.8 & 12.3 & 34.4 & 18.2 & 37.3 & \textbf{+22.7} \\
ReasonEval-7B & 19.8 & 97.4 & 32.9 & 27.4 & 82.5 & 41.2 & 22.7 & 64.6 & 33.6 & 21.5 & 72.6 & 33.1 & 35.2 & \textbf{+24.8} \\
Qwen2.5-Math-7B-Math-Shepherd* & 46.4 & 95.9 & 62.4 & 18.9 & 96.6 & 31.6 & 7.4 & 93.8 & 13.7 & 4.0 & 95.6 & 7.7 & 28.9 & \textbf{+31.1} \\
Qwen2.5-Math-7B-PRM800K & 52.8 & 94.7 & 67.8 & 47.8 & 91.2 & 62.7 & 33.7 & 85.2 & 48.3 & 27.1 & 87.2 & 41.3 & 55.0 & \textbf{+5.0} \\
OVM-Qwen3-4B & 4.8 & 56.5 & 8.9 & 37.7 & 85.5 & 52.3 & 31.0 & 79.9 & 44.7 & 23.3 & 80.5 & 36.2 & 35.5 & \textbf{+24.5} \\
PQM-DeepSeek-zeta4-7B & 31.9 & 40.9 & 35.8 & 22.2 & 19.7 & 20.9 & 8.9 & 6.2 & 7.3 & 11.1 & 15.8 & 13.0 & 19.2 & \textbf{+40.8} \\
\addlinespace[1pt]
\textbf{SPRM-Qwen3-4B} & 55.5 & 94.8 & 70.1 & 52.4 & 76.0 & 62.8 & 48.1 & 60.0 & 53.2 & 48.6 & 60.2 & 53.8 & \textbf{60.0} & - \\
\bottomrule
\end{tabular}%
}
\end{table*}

\begin{figure*}[t]
    \centering
    \includegraphics[width=\linewidth]{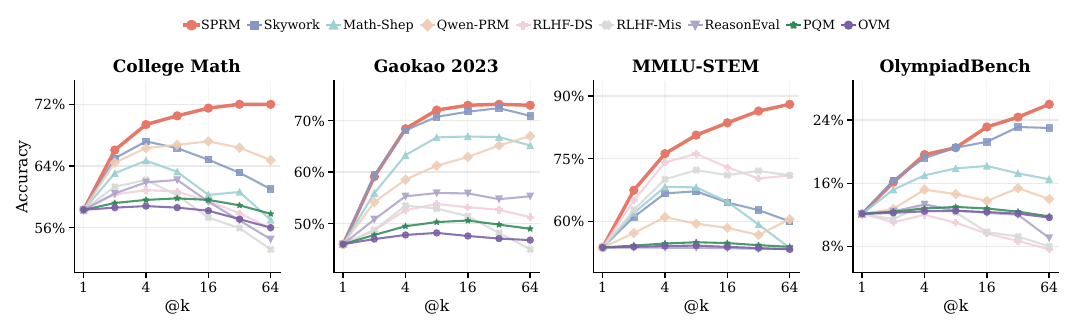}
    \caption{Best-of-N inference with different PRMs on four representative
    datasets.
    Each subplot reports accuracy as a function of \(k\) on one dataset.}
    \label{fig:bon_prm_grid}
\end{figure*}

\begin{table*}[t!]
\centering
\footnotesize
\tabtight
\caption{PRMBench results.
\(\Delta = \mathrm{SPRM} - \mathrm{baseline}\) for the Overall score;
positive values indicate higher scores for SPRM.
Data for models marked with an asterisk (\(*\)) are sourced from
\citep{DBLP:prmbench}.}
\label{tab:prmbench_types_en}
\resizebox{0.97\textwidth}{!}{%
\begin{tabular}{@{}l*{12}{C{2.75em}}C{4.0em}C{2.9em}@{}}
\toprule
\multirow{2}{*}{\textbf{Model}} & \multicolumn{3}{c}{\textbf{Simplicity}} & \multicolumn{5}{c}{\textbf{Soundness}} & \multicolumn{4}{c}{\textbf{Sensitivity}} & \multirow{2}{*}{\textbf{Overall}} & \multirow{2}{*}{\(\mathbf{\Delta}\)} \\
\cmidrule(lr){2-4}\cmidrule(lr){5-9}\cmidrule(lr){10-13}
 & NR & NCL & Avg & ES & SC & DC & CI & Avg & PS & DR & MS & Avg & & \\
\midrule
\tabgroup{15}{Closed-source judges}
GPT-4o* & 57.0 & 62.4 & 59.7 & 72.0 & 69.7 & 70.7 & 71.1 & 70.9 & 62.5 & 65.7 & 99.2 & 75.8 & 66.8 & -4.4 \\
o1-mini* & 65.6 & 63.7 & 64.6 & 74.5 & 67.7 & 73.8 & 72.3 & 72.1 & 61.8 & 64.8 & 100.0 & 75.5 & 68.8 & -6.4 \\
\midrule
\tabgroup{15}{PRMs}
Math-Shepherd-7B & 42.3 & 47.3 & 44.8 & 50.1 & 42.3 & 41.2 & 48.0 & 45.4 & 43.1 & 49.6 & 90.1 & 60.9 & 46.6 & \textbf{+15.8} \\
RLHFLOW-Mistral-8B & 46.3 & 47.1 & 46.7 & 56.5 & 54.8 & 54.0 & 63.9 & 57.3 & 51.0 & 55.9 & 98.2 & 68.4 & 54.2 & \textbf{+8.2} \\
RLHFLOW-DeepSeek-8B & 46.4 & 48.5 & 47.5 & 55.0 & 54.5 & 53.1 & 65.6 & 57.1 & 48.8 & 55.0 & 99.7 & 67.8 & 53.8 & \textbf{+8.6} \\
Skywork-PRM-7B & 40.5 & 48.1 & 44.3 & 45.5 & 36.8 & 38.0 & 43.3 & 40.9 & 42.7 & 45.5 & 93.3 & 60.5 & 43.3 & \textbf{+19.1} \\
ReasonEval-7B & 49.2 & 57.1 & 53.2 & 62.1 & 66.0 & 61.5 & 65.8 & 63.9 & 55.6 & 58.0 & 99.1 & 70.9 & 59.8 & \textbf{+2.6} \\
Qwen2.5-Math-7B-PRM800K & 48.6 & 47.8 & 48.2 & 62.1 & 59.4 & 58.7 & 68.5 & 62.2 & 52.9 & 64.0 & 99.8 & 72.2 & 59.1 & \textbf{+3.3} \\
OVM-Qwen3-4B & 50.0 & 43.4 & 46.7 & 45.5 & 48.7 & 46.4 & 47.3 & 47.0 & 46.3 & 47.7 & 95.2 & 63.1 & 47.2 & \textbf{+15.2} \\
PQM-DeepSeek-zeta4-7B & 17.7 & 25.7 & 21.7 & 22.9 & 16.1 & 16.6 & 18.4 & 18.5 & 21.6 & 20.8 & 43.7 & 28.7 & 20.3 & \textbf{+42.1} \\
\addlinespace[1pt]
\textbf{SPRM-Qwen3-4B} & 52.5 & 64.9 & 58.7 & 67.5 & 59.8 & 58.6 & 65.1 & 62.8 & 60.0 & 63.9 & 99.1 & 74.3 & \textbf{62.4} & - \\
\bottomrule
\end{tabular}%
}
\end{table*}
\paragraph{SPRM Addresses the Three Targeted Evaluation Properties}
\textbf{First-error localization.}
On ProcessBench, SPRM achieves the best average F1 of 60.0, improving over
Math-Shepherd-7B and RLHFlow by 28.1--32.8 points and over the same-backbone
OVM-Qwen3-4B by 24.5 points.
This gap suggests that rollout labels and outcome broadcasting still blur the
marginal cause of failure, while SPRM better separates the first harmful step
from its surrounding context without simply accepting more trajectories as
correct.
\textbf{Marginal-contribution preferences.}
On PRMBench, SPRM obtains the best overall score among open PRMs at 62.4, with
strong \textit{Simplicity} and \textit{Sensitivity} scores of 58.7 and 74.3;
its NCL score of 64.9 and PS score of 60.0 show that it better down-weights
redundant reasoning and penalizes harmful perturbations.
\textbf{Reasoning-time selection.}
\Cref{fig:bon_prm_grid} shows that SPRM scales reliably as \(k\) increases to
64, reaching 72.0 on CollegeMath, 73.0 on Gaokao2023en, 97.0 on GSM8K, and 90.0
on MATH500.
By contrast, OVM drops on CollegeMath and MATH500 as \(k\) grows from 2 to 64,
while Math-Shepherd largely saturates, indicating that marginal-contribution
supervision gives a more stable selection signal under larger search pools.

\section{Analysis}
\label{sec:analysis}
This section asks four \textbf{R}esearch \textbf{Q}uestions (\textbf{RQ})
about the mechanisms and cross-domain applicability of the constructed
supervision.
RQ1--RQ3 isolate the roles of SPRM's supervision components while keeping the
backbone, training distribution, and optimization setup fixed, so performance
differences primarily reflect the supervision signal itself.
These three experiments use ProcessBench and report \textit{Err}, \textit{Cor},
and F1, as summarized in \Cref{tab:ablation_processbench}.
RQ4 tests transfer of the supervision framework to code generation through
Best-of-N selection accuracy.
The four RQs are:

\textbf{RQ1.} \textit{Can SVI provide more learnable step credit than temporal
value differences?}

\textbf{RQ2.} \textit{Why choose \(R^{\mathrm{count}}\)-consistency instead
of LLM-consistency?}

\textbf{RQ3.} \textit{Can cross-trajectory inherent score reduce context-induced credit fluctuations?}

\textbf{RQ4.} \textit{Can SPRM's outcome-to-step supervision transfer beyond
mathematical reasoning?}

\paragraph{RQ1: Can SVI provide more learnable step credit than temporal
value differences?}
To answer RQ1, we construct three PRM training signals in
\Cref{tab:ablation_processbench}: delta-value labels, SVI labels without
\textit{question\_end}, and full SVI labels with \textit{question\_end}.
This ablation is designed to test whether the supervision quality improves when
step credit is derived from one trajectory-level predictor, and whether the
reasoning-start boundary should participate in the allocation.

The motivation is that delta-value labels propagate prefix-value errors.
If \(\hat v(P_t)=v^\star(P_t)+\epsilon_t\), then the sign of the delta label is
stable only when
\(
|\epsilon_t-\epsilon_{t-1}| < |v^\star(P_t)-v^\star(P_{t-1})|
\);
small prefix errors can therefore flip labels when the true marginal gap is
small.
SVI does not remove value-estimation error, but changes it from a difference of
two prefix residuals into the path-integrated attribution of one trajectory
residual.
Let \(v_{\theta}^{\mathrm{traj}}=v^{\star,\mathrm{traj}}+e\).
By completeness \citep{sundararajan2017axiomatic}, the residual in
\(v_{\theta}^{\mathrm{traj}}(X)-v_{\theta}^{\mathrm{traj}}(B)\) is assigned over
the encoded participants used in the allocation.
This explains why full SVI improves average F1 from 35.5 to 50.2.
It also explains the boundary ablation: the reasoning start may itself carry
value mismatch, and excluding \textit{question\_end} forces this error onto the
first reasoning step, reducing F1 to 36.1.
Thus, RQ1 is answered positively: SVI gives more learnable step supervision by
avoiding prefix-error differencing, provided that the allocation includes the
reasoning-start participant.

\begin{table*}[t]
    \centering
    \footnotesize
    \tabtight
    \caption{ProcessBench ablations isolating SPRM's three design modules:
    SVI step-credit source, model-counterfactual consistency mechanism,
    cross-trajectory inherent score, and the SVI/CTI fusion weight \(\lambda\),
    to assess their individual contributions.
    Gray shading marks the configurations used in our method.}
    \label{tab:ablation_processbench}
    \resizebox{\textwidth}{!}{%
    \begin{tabular}{@{}p{0.095\textwidth}p{0.19\textwidth}*{3}{C{0.038\textwidth}}@{\hspace{0.9em}}!{\color{black!25}\vrule width 0.35pt}@{\hspace{0.25em}}!{\color{black!25}\vrule width 0.35pt}@{\hspace{0.9em}}p{0.095\textwidth}p{0.19\textwidth}*{3}{C{0.038\textwidth}}@{}}
    \toprule
    \textbf{Component} & \textbf{Setting} & \textbf{Err} & \textbf{Cor} & \textbf{F1}
    & \textbf{Component} & \textbf{Setting} & \textbf{Err} & \textbf{Cor} & \textbf{F1} \\
    \midrule
    \multirow{4}{0.095\textwidth}{\raggedright\textit{Step-credit source}}
    & \multirow{4}{0.19\textwidth}{\parbox[c][3.1\baselineskip][s]{0.19\textwidth}{%
    \raggedright
    Delta value\vfill
    SVI w/o {\itshape question\_end}\vfill
    {\setlength{\fboxsep}{0pt}\makebox[\linewidth][l]{%
    \hspace*{-\tabcolsep}\colorbox{TableBandGray}{%
    \hspace{\tabcolsep}\makebox[\linewidth][l]{\strut SVI}%
    \hspace{\tabcolsep}}}}}}
    & \multirow{4}{0.038\textwidth}{\parbox[c][3.1\baselineskip][s]{0.038\textwidth}{%
    \centering
    24.2\vfill
    31.4\vfill
    {\setlength{\fboxsep}{0pt}\makebox[\linewidth][l]{%
    \hspace*{-\tabcolsep}\colorbox{TableBandGray}{%
    \hspace{\tabcolsep}\makebox[\linewidth][c]{\strut \textbf{37.7}}%
    \hspace{\tabcolsep}}}}}}
    & \multirow{4}{0.038\textwidth}{\parbox[c][3.1\baselineskip][s]{0.038\textwidth}{%
    \centering
    75.6\vfill
    42.6\vfill
    {\setlength{\fboxsep}{0pt}\makebox[\linewidth][l]{%
    \hspace*{-\tabcolsep}\colorbox{TableBandGray}{%
    \hspace{\tabcolsep}\makebox[\linewidth][c]{\strut \textbf{76.5}}%
    \hspace{\tabcolsep}}}}}}
    & \multirow{4}{0.038\textwidth}{\parbox[c][3.1\baselineskip][s]{0.038\textwidth}{%
    \centering
    35.5\vfill
    36.1\vfill
    {\setlength{\fboxsep}{0pt}\makebox[\linewidth][l]{%
    \hspace*{-\tabcolsep}\colorbox{TableBandGray}{%
    \hspace{\tabcolsep}\makebox[\linewidth][c]{\strut \textbf{50.2}}%
    \hspace{\tabcolsep}}}}}}
    & \multirow{4}{0.095\textwidth}{\raggedright\textit{CTI score}}
    & SVI only & 42.7 & \textbf{79.0} & 55.4 \\
    & & & & & & SVI + random & 43.8 & 74.6 & 55.2 \\
    & & & & & & CTI only & 26.0 & 46.9 & 33.4 \\
    & & & & & & \cellcolor{TableBandGray} SVI + CTI
    & \cellcolor{TableBandGray} \textbf{51.2}
    & \cellcolor{TableBandGray} 73.3
    & \cellcolor{TableBandGray} \textbf{60.0} \\
    \midrule
    \multirow{5}{0.095\textwidth}{\raggedright\shortstack[l]{\textit{Consistency}\\\textit{Mechanism}}}
    & No consistency & 37.7 & 76.5 & 50.2
    & \multirow{5}{0.095\textwidth}{\centering\(\lambda\), (\(1 - \lambda\))} & \((1.00, 0.00)\) & 42.7 & 79.0 & 55.4 \\
    & & & &
    & & \((0.75, 0.25)\) & 40.9 & \textbf{83.1} & 54.7 \\
    & LLM-consistency & \textbf{47.3} & 67.0 & \textbf{55.4}
    & & \cellcolor{TableBandGray} \((0.50, 0.50)\)
    & \cellcolor{TableBandGray} \textbf{51.2}
    & \cellcolor{TableBandGray} 73.3
    & \cellcolor{TableBandGray} \textbf{60.0} \\
    & & & &
    & & \((0.25, 0.75)\) & 34.3 & 60.6 & 43.8 \\
    & \cellcolor{TableBandGray} \(R^{\mathrm{count}}\)-consistency
    & \cellcolor{TableBandGray} 42.7
    & \cellcolor{TableBandGray} \textbf{79.0}
    & \cellcolor{TableBandGray} \textbf{55.4}
    & & \((0.00, 1.00)\) & 26.0 & 46.9 & 33.4 \\
    \bottomrule
    \end{tabular}%
    }
    \par\vspace{8pt}

\centering
\footnotesize
\tabtight
\caption{Cross-domain transfer to code generation: selection accuracy (\%)
on LiveCodeBench. Entries report mean $\pm$ standard deviation over three
candidate-sampling seeds; BoN@32 uses the entire fixed pool and thus has zero
standard deviation. Base is model accuracy before training; Pass@$K$ is the
upper bound indicating whether any candidate passes all tests.
Boldface marks the best mean excluding this upper bound.}
\label{tab:code_bon_full}

\begin{tabular*}{\textwidth}{@{\extracolsep{\fill}}lcccc@{}}
\toprule
\textbf{Method} & \textbf{BoN@4} & \textbf{BoN@8}
    & \textbf{BoN@16} & \textbf{BoN@32} \\
\midrule
Pass@$K$ (upper bound)
    & 49.1 $\pm$ 0.3
    & 52.3 $\pm$ 0.1
    & 53.0 $\pm$ 0.6
    & 57.2 $\pm$ 0.0 \\
\midrule
Base (before training)
    & 40.0 $\pm$ 0.5
    & 40.9 $\pm$ 0.7
    & 41.7 $\pm$ 0.2
    & 46.8 $\pm$ 0.0 \\
OVM
    & 43.5 $\pm$ 0.3
    & 44.0 $\pm$ 0.8
    & 43.8 $\pm$ 1.0
    & 48.6 $\pm$ 0.0 \\
\textbf{SPRM}
    & \textbf{43.7} $\pm$ \textbf{0.8}
    & \textbf{46.2} $\pm$ \textbf{0.7}
    & \textbf{47.9} $\pm$ \textbf{1.1}
    & \textbf{52.1} $\pm$ \textbf{0.0} \\
\bottomrule
\end{tabular*}
\end{table*}

\paragraph{RQ2: Why choose \(R^{\mathrm{count}}\)-consistency instead of
LLM-consistency?}
To answer RQ2, we design a consistency ablation that trains PRMs with three
filters on SVI credit: no consistency filter, an LLM-consistency filter using
DeepSeek-V3.2 \citep{DBLP:deepseek_v3_2} (2026-03-24), and our model-counterfactual
consistency filter, denoted by \(R^{\mathrm{count}}\).
This experiment directly tests whether an external strong LLM judge is a better
consistency source than the trajectory-value-based interaction score.
LLM-consistency gives the highest \textit{Err} score, 47.3, indicating stricter
error correction, which is expected from a strong model with substantial logical
checking ability.
However, it lowers \textit{Cor} to 67.0, consistent with observations that LLMs
can be unstable when producing long structured arrays
\citep{islam2026numericalinstabilitychaosquantifying}.
By contrast, \(R^{\mathrm{count}}\)-consistency preserves the strongest
\textit{Cor} score of 79.0 while matching the best F1 of 55.4 and improving over
the no-consistency F1 of 50.2.
This follows from its design: \(R_{i,(t,u)}^{\mathrm{int}}\) in
\Cref{eq:counterfactual_effects} measures whether a step supports local
interactions with previous steps, and \Cref{eq:pairwise_counterfactual} only
aggregates positive interaction terms.
Thus, \(R^{\mathrm{count}}\) emphasizes steps that participate in a fluent
reasoning chain rather than aggressively penalizing every local imperfection,
which helps preserve truly useful steps for PRM learning.

\paragraph{RQ3: Can cross-trajectory inherent score reduce context-induced credit fluctuations?}
To answer RQ3, we analyze whether cross-trajectory evidence can stabilize
trajectory-specific credit.
SVI is trajectory-conditioned: the same step pattern may receive different
credit when it appears after a flawed prefix, within an easy problem, or near a
decisive later correction.
CTI addresses this variance by estimating a non-parametric conditional-success
tendency:
\(
\delta^{\mathrm{cti}}(x)
\approx
P(\mathrm{success}\mid \mathrm{similar\ step\ appears})
-
P(\mathrm{success}).
\)
SVI asks how the current step contributes within its current trajectory, whereas
CTI asks whether semantically similar steps are statistically associated with
successful trajectories in the data distribution.

We therefore design the CTI ablations to separate true cross-trajectory
complementarity from generic smoothing: ``SVI only'' uses trajectory-internal
credit, ``CTI only'' uses the inherent score alone, ``SVI + CTI'' combines the
two, and ``SVI + random'' replaces semantic neighbors with random ones.
``CTI only'' is weak because it lacks the current trajectory's causal context,
but adding CTI to SVI improves average F1 from 55.4 to 60.0.
Replacing retrieved neighbors with random ones drops F1 to 55.2, showing that
the gain comes from semantic cross-trajectory evidence rather than arbitrary
averaging.
The head-mixing rows give the same conclusion: too much CTI weight removes
trajectory-internal context, while the balanced setting gives the best F1.
Together with the trajectory-mean \(\delta^{\mathrm{cti}}\) AUC of 0.856 against
final reward in \Cref{sec:appendix_cti_score_analysis}, these results support
CTI as a stabilizer for SVI rather than a replacement for trajectory-specific
credit.

\paragraph{RQ4: Can SPRM's outcome-to-step supervision transfer beyond
mathematical reasoning?}
To answer RQ4, we examine whether outcome-based step credit also benefits
code generation by retraining SPRM on code trajectories with the same PRM
training configuration as in the mathematics experiments.
Qwen3-4B-Instruct-2507 generates the BigCodeBench
\cite{DBLP:BigCodeBench} training trajectories, segmented into function-level
steps following FunPRM \cite{DBLP:FunPRM}.
The dataset contains 4,697 trajectories and 15,040 steps; only programs passing
all unit tests receive positive terminal labels, while partially passing and
failing programs receive negative labels.
We train a same-backbone OVM on the same data by broadcasting terminal labels
to all steps.
This comparison tests whether allocating outcome feedback to individual steps
is more effective than assigning the same terminal label to every step.
For evaluation, the same generator produces a fixed pool of 32 candidates per
problem on the LiveCodeBench v6 \cite{DBLP:LiveCodeBench} slice starting on
2025-02-01, retaining 134 executable problems.
We sample and rerank \(K \in \{4,8,16,32\}\) candidates per problem and report
mean accuracy and standard deviation over three candidate-sampling seeds.

\Cref{tab:code_bon_full} shows that SPRM has higher mean accuracy than OVM
at all four budgets: the gap is small at BoN@4, and increases to 2.2, 4.1, and
3.5 percentage points at BoN@8, BoN@16, and BoN@32, respectively.
At BoN@32, SPRM reaches 52.1\% compared with OVM's 48.6\%.
The gains over terminal-label broadcasting suggest that SPRM can extract
useful step-level supervision in code generation, providing evidence of its
transfer potential beyond mathematical reasoning under reliable terminal
verification.
Extending this framework to open-ended tasks or noisier reward settings is a
natural direction for future work.

\section{Conclusion}
We presented SPRM, an outcome-only framework that derives process reward
supervision from sparse terminal feedback via an ordered cooperative game.
By combining Aumann-Shapley-style attribution, model-counterfactual consistency, and
cross-trajectory inherent scoring, SPRM produces scalable step-level supervision
without strong external supervision.
Experiments on verifiable reasoning tasks show consistent gains in first-error
localization, trajectory-level evaluation, and Best-of-N reranking.
A key limitation is that the current method assumes verifiable outcome rewards,
and CTI depends on embedding quality for reliable semantic retrieval.
Future work will extend this perspective to reward modeling for tasks where
answers cannot be reliably verified.

\section*{Acknowledgments}

This work is supported
by the National Key R\&D Program of China (2026YFE0199800),
Science and Technology Development Fund of Macao S.A.R (FDCT)
under 0123-2022-AFJ, 0081-2022-A2, and 0074-2025-AMJ,
National Natural Science Foundation of China (62376263),
Natural Science Foundation of Guangdong (2024A1515030209),
and Shenzhen Science and Technology Innovation Commission
(JCYJ20230807140507015).

\bibliographystyle{plainnat}
\bibliography{references}
\newpage
\appendix
\section{Assumptions and Theoretical Analysis}
\label{sec:appendix}

\paragraph{Appendix roadmap}
The appendix is organized into three parts.
\Cref{sec:appendix} gives the assumptions and theoretical analysis behind the
two lemmas used in the main text.
It first fixes notation, states the assumptions under which all proofs are
interpreted, compares temporal credit assignment with classical Shapley
allocation, and then proves the two lemmas used in the main text.
\Cref{sec:appendix_additional_experiments} collects supplementary empirical
material, including baseline supervision comparisons, evaluation metrics,
compute-cost comparisons, design studies, attribution diagnostics, CTI
diagnostics, and complete Best-of-N results.
\Cref{sec:appendix_impl_details} gives implementation and environment details
for reproducing the reported runs.

\subsection{Notation}
\label{sec:appendix_notation}

\begin{table}[h!]
\centering
\caption{Notation used throughout the appendix, aligned with
\Cref{sec:formulation} and \Cref{sec:method_step}.}
\label{tab:appendix_notation}
\small
\begin{tabular}{p{0.28\linewidth}p{0.62\linewidth}}
\toprule
Symbol & Meaning \\
\midrule
\(\mathcal{D}\) &
Outcome-supervised trajectory dataset. \\
\(\tau_i\) &
Trajectory containing question \(q_i\) and \(T_i\) reasoning steps. \\
\(r_i\) &
Terminal ground-truth reward of trajectory \(\tau_i\). \\
\(\mathcal{N}_i\) &
Participant set in the discrete reasoning game. \\
\(s_{i,t}\) &
Step \(t\) in trajectory \(\tau_i\), treated as a participant in the discrete
reasoning game. \\
\(\Pi_{\mathrm{adm}}(\mathcal{N}_i)\) &
Set of temporally admissible permutations of the participant set
\(\mathcal{N}_i\). \\
\(\pi^\star\) &
Unique admissible temporal permutation
\((s_{i,1},s_{i,2},\ldots,s_{i,T_i})\) under Assumption A1. \\
\(\mathrm{Pre}_{\pi}(s_{i,t})\) &
Set of steps that precede \(s_{i,t}\) under permutation \(\pi\). \\
\(P_{i,t}\) &
Prefix coalition up to step \(t\). \\
\(v\) &
Generic value function that maps a coalition, prefix, or encoded trajectory to a
scalar value. \\
\(\phi_{i,t}^{\mathrm{ord}}\) &
Ordered Shapley-style credit assigned to step \(t\) under the admissible temporal
order. \\
\(X_i\) &
Encoded question and step representations. \\
\(x_{i,0}\), \(x_{i,t}\) &
Question representation and step-\(t\) representation, respectively. \\
\(b\), \(B_i\) &
Semantically empty baseline feature and baseline sequence. \\
\(\gamma_i\) &
Straight-line path function
\(\gamma_i(\alpha)=B_i+\alpha(X_i-B_i)\). \\
\(\delta_{i,t}^{\mathrm{cre}}\) &
SVI credit of step \(t\), as defined in \Cref{eq:delta_step}. \\
\(R_{i,t}^{\mathrm{count}}\) &
Model-counterfactual consistency score used to mask step-credit labels. \\
\(\delta_{i,t}^{\mathrm{cti}}\) &
Cross-trajectory inherent score of step \(t\). \\
\(\psi\) &
Generic allocation rule for the continuous attribution problem. \\
\bottomrule
\end{tabular}
\end{table}

\subsection{Assumptions}
\label{sec:appendix_assumptions}

All theoretical statements below use the notation in \Cref{tab:appendix_notation}
and are interpreted under the following assumptions.
\textbf{A1. Temporal order.}
For each reasoning trajectory, the only admissible causal order is
\(1\prec2\prec\cdots\prec T_i\); later steps are interpreted after their
observed prefixes and are not freely exchangeable with earlier steps.
\textbf{A2. Fixed attribution problem.}
For SVI, all attribution claims are made with respect to a fixed trajectory value
function \(v\) and a fixed baseline \(B_i\).
The path function is the straight-line map
\(\gamma_i(\alpha)=B_i+\alpha(X_i-B_i)\).
\textbf{A3. Regularity.}
The scalar path function \(\alpha\mapsto v(\gamma_i(\alpha))\) is absolutely
continuous on \([0,1]\), and each partial gradient
\(\nabla_{x_{i,t}}v(\gamma_i(\alpha))\) exists almost everywhere and is
integrable along this path.
For linearity comparisons, the compared value functions use the same baseline.

\subsection{Temporal Shapley Comparison}
\label{sec:appendix_shapley_axioms}

This section compares the SPRM step-credit term in \Cref{eq:delta_step} with
classical Shapley axioms.
The goal is not to claim that temporal credit assignment exactly matches the
classical Shapley value, because reasoning steps are causally ordered and not
freely exchangeable.
Instead, we clarify which Shapley-inspired properties SPRM satisfies, where the
temporal setting departs from the classical axioms, and what fixed value
function, fixed-baseline, and fixed-path constraints are required.
The discussion applies only to the attribution signal
\(\delta_{i,t}^{\mathrm{cre}}\), not to the counterfactual filter, CTI score, or
binarized PRM labels.

\paragraph{Classical Shapley axioms}
For a finite cooperative game with participant set \(N\), value function
\(v:2^N\rightarrow\mathbb{R}\), and allocation rule
\(\phi\), which maps a value function \(v\) to one real-valued credit
\(\phi(v)=(\phi_j(v))_{j\in N}\in\mathbb{R}^{|N|}\) for each participant, the
Shapley value is the unique allocation satisfying the four axioms in
\Cref{tab:classical_shapley_axioms}
\citep{shapley1953value}.

These axioms determine a unique notion of marginal-contribution credit:
the classical Shapley value averages each participant's marginal contribution
over all permutations of the participants.
For continuous feature spaces, \citet{sundararajan2017axiomatic} show that
gradient integration is the unique attribution rule satisfying
Completeness, Sensitivity, Linearity, and Symmetry-Preserving.
These four conditions respectively correspond to the four classical Shapley
axioms in \Cref{tab:classical_shapley_axioms}: efficiency, dummy participant,
additivity, and symmetry.

\paragraph{Temporal constraints without exchangeability}
The question in temporally ordered reasoning is whether we can still obtain a
Shapley-like credit rule when exchangeability is unavailable, and what is lost
relative to the classical characterization.
Reasoning trajectories differ from unordered cooperative games because a later
step is causally interpreted after its prefix, and permuting steps generally
changes the reasoning process.
Thus the classical symmetry axiom, and likewise Symmetry-Preserving in the
continuous setting, cannot be matched literally.
To stay as close as possible to Shapley's fair-allocation principle, we define
the trajectory game over the admissible prefix chain
\(
P_{i,0}\subset P_{i,1}\subset\cdots\subset P_{i,T_i}
\).
On this chain, there is only one admissible temporal order.
Thus the Shapley average over permutations collapses to the ordered marginal
contribution \(v(P_{i,t})-v(P_{i,t-1})\), as stated in
\Cref{lem:temporal_shapley_prefix}.
The temporal setting therefore replaces the symmetry-based characterization
with three ordered properties that preserve the parts of Shapley most relevant
to causal credit assignment:
\textit{temporal efficiency}, inherited from efficiency, requires all payoff
difference to be allocated over the ordered positions;
\textit{prefix dummy}, inherited from the dummy-participant axiom, gives zero credit
to a step whose addition to its prefix changes no value; and
\textit{linearity}, inherited from additivity, requires allocations to commute
with linear combinations of value functions.
The gap from classical Shapley is precisely symmetry:
we do not assert exchangeability among reasoning steps, but instead use the
prefix-order constraint as the admissible causal structure.

\paragraph{Continuous SPRM setting}
SPRM then moves from discrete prefix values to continuous feature
attribution.
Under Assumptions A2--A3, \Cref{lem:svi_shapley_like_properties} shows that the
SPRM credit in \Cref{eq:delta_step} satisfies the continuous counterparts of
efficiency, dummy, and linearity along the path function \(\gamma_i\).
This comparison makes the limitation explicit:
SPRM does not satisfy classical Shapley symmetry, because temporal reasoning
steps are not exchangeable under the causal order.
Thus this appendix positions SPRM as a Shapley-inspired credit assignment rule
under temporal-causal constraints, rather than as a full classical Shapley
value.

\subsection{Lemma Proofs}
\label{sec:appendix_step_credit_proofs}

\subsubsection{Proof of the Ordered Prefix-Difference Lemma}

\begin{proof}[Proof of \Cref{lem:temporal_shapley_prefix}]
Let \(\Pi_{\mathrm{adm}}(\mathcal{N}_i)\) denote the set of temporally admissible
permutations of the step set \(\mathcal{N}_i=\{s_{i,1},\ldots,s_{i,T_i}\}\).
The admissible-order Shapley-style credit is the marginal-contribution average
restricted to this set:
\begin{equation}
\label{eq:appendix_ordered_shapley_average}
\phi_{i,t}^{\mathrm{ord}}
:=
\frac{1}{|\Pi_{\mathrm{adm}}(\mathcal{N}_i)|}
\sum_{\pi\in\Pi_{\mathrm{adm}}(\mathcal{N}_i)}
\Big[
v\bigl(\mathrm{Pre}_{\pi}(s_{i,t})\cup\{s_{i,t}\}\bigr)
-
v\bigl(\mathrm{Pre}_{\pi}(s_{i,t})\bigr)
\Big].
\end{equation}
By Assumption A1, the only admissible order is
\begin{equation}
\pi^\star=(s_{i,1},s_{i,2},\ldots,s_{i,T_i}),
\qquad
\Pi_{\mathrm{adm}}(\mathcal{N}_i)=\{\pi^\star\}.
\end{equation}
Thus \(|\Pi_{\mathrm{adm}}(\mathcal{N}_i)|=1\), and
\Cref{eq:appendix_ordered_shapley_average} becomes
\begin{equation}
\phi_{i,t}^{\mathrm{ord}}
=
v\bigl(\mathrm{Pre}_{\pi^\star}(s_{i,t})\cup\{s_{i,t}\}\bigr)
-
v\bigl(\mathrm{Pre}_{\pi^\star}(s_{i,t})\bigr).
\end{equation}
Under \(\pi^\star\), the predecessors of step \(s_{i,t}\) are exactly the
previous prefix:
\begin{equation}
\mathrm{Pre}_{\pi^\star}(s_{i,t})
=
\{s_{i,1},\ldots,s_{i,t-1}\}
=
P_{i,t-1}.
\end{equation}
Moreover,
\begin{equation}
\mathrm{Pre}_{\pi^\star}(s_{i,t})\cup\{s_{i,t}\}
=
\{s_{i,1},\ldots,s_{i,t}\}
=
P_{i,t}.
\end{equation}
Substituting these two identities yields
\begin{equation}
\phi_{i,t}^{\mathrm{ord}}
=
v(P_{i,t})-v(P_{i,t-1}),
\end{equation}
which proves the claim.
\end{proof}

\subsubsection{Proof of the SVI Shapley-Like Lemma}

\begin{proof}[Proof of \Cref{lem:svi_shapley_like_properties}]
The three Shapley-like properties considered here are discussed and motivated
in \Cref{sec:appendix_shapley_axioms}.
We prove that temporal efficiency, prefix dummy, and linearity survive under
the fixed trajectory value function and baseline, with the path function
\(\gamma_i\), in Assumptions A2--A3.
The SVI credit is
\begin{equation}
\delta_{i,t}^{\mathrm{cre}}
:=
(x_{i,t}-b)^\top
\int_0^1
\nabla_{x_{i,t}}v(\gamma_i(\alpha))
\,d\alpha .
\end{equation}

\paragraph{Temporal efficiency}
By Assumption A3 and the fundamental theorem of calculus along \(\gamma_i\),
\begin{equation}
v(X_i)-v(B_i)
=
\int_0^1
\nabla v(\gamma_i(\alpha))^\top (X_i-B_i)
\,d\alpha .
\end{equation}
Decomposing the inner product over encoded positions gives
\begin{equation}
\sum_{t=0}^{T_i}\delta_{i,t}^{\mathrm{cre}}
=
v(X_i)-v(B_i).
\end{equation}
If only reasoning steps are summed, the question representation is excluded:
\begin{equation}
\sum_{t=1}^{T_i}\delta_{i,t}^{\mathrm{cre}}
=
v(X_i)-v(B_i)-\delta_{i,0}^{\mathrm{cre}}.
\end{equation}
This is the ordered analogue of efficiency for the fixed continuous attribution
problem.

\paragraph{Prefix dummy}
If the encoded step block has zero marginal effect along the path, then it
receives zero credit.
Formally, if
\(
\nabla_{x_{i,t}}v(\gamma_i(\alpha))=0
\)
for almost all \(\alpha\in[0,1]\), then
\(\delta_{i,t}^{\mathrm{cre}}=0\).
This statement does not require any special value of \(v(B_i)\); the baseline
value only sets the reference level in the completeness equation.

\paragraph{Linearity}
For two value functions \(v\) and \(w\) evaluated with the same baseline and
path, let \(h=av+cw\).
Then
\(\nabla h=a\nabla v+c\nabla w\), and therefore
\begin{equation}
\delta_{i,t}^{\mathrm{cre}}(h)
:=
a\delta_{i,t}^{\mathrm{cre}}(v)
+
c\delta_{i,t}^{\mathrm{cre}}(w).
\end{equation}
Thus \(\delta_{i,t}^{\mathrm{cre}}\) satisfies the continuous counterparts of
efficiency, dummy, and linearity under temporal-causal constraints
\citep{sundararajan2017axiomatic,pmlr-v119-sundararajan20b}.
\end{proof}

\clearpage

\section{Additional Experiments}
\label{sec:appendix_additional_experiments}

This appendix section collects supplementary material for the empirical part of
the paper.
We first compare baseline supervision regimes, define the evaluation metrics,
then report the compute-cost comparison, additional analyses, and the full
Best-of-N tables omitted from the main text.

\subsection{Baseline Supervision Regimes}
\label{sec:appendix_baseline_regimes}

\begin{table*}[h]
    \centering
    \small
    \setlength{\tabcolsep}{4.5pt}%
    \renewcommand{\arraystretch}{1.12}%
    \caption{Baseline supervision regimes.
    ``Auto.'' marks automatically constructed process supervision;
    ``No Ext.'' marks no external annotators, judges, or curated intermediate
    labels; ``Marginal'' marks whether step supervision accounts for marginal
    contribution.}
    \label{tab:baseline_supervision_compare}
    \begin{tabular*}{0.98\textwidth}{@{\extracolsep{\fill}}
    >{\raggedright\arraybackslash}p{0.22\textwidth}
    >{\raggedright\arraybackslash}p{0.40\textwidth}
    C{0.075\textwidth}
    C{0.075\textwidth}
    C{0.09\textwidth}@{}}
    \toprule
    \textbf{Paradigm} & \textbf{Representative models} & \textbf{Auto.} & \textbf{No Ext.} & \textbf{Marg. contrib.} \\
    \midrule
    Human labels & Qwen2.5-Math-7B-PRM800K~\citep{zhang-etal-2025-lessons}; ReasonEval~\citep{DBLP:ReasonEval} & \XSolidBrunsh & \XSolidBrunsh & \XSolidBrunsh \\
    MC labels & Math-Shepherd~\citep{wang2024math}; RLHFlow~\citep{dong2024rlhf} & \checkmark & \XSolidBrunsh & \XSolidBrunsh \\
    Outcome broadcast & OVM-Qwen3-4B~\citep{ovm:conf/naacl/YuGW24} & \checkmark & \checkmark & \XSolidBrunsh \\
    First-error boundary & PQM-DeepSeek-zeta4-7B~\citep{DBLP:conf/iclr/LiL25a} & \XSolidBrunsh & \checkmark & \XSolidBrunsh \\
    Curated process & Skywork-PRM-7B \citep{skywork} & - & - & \XSolidBrunsh \\
    \textbf{Outcome-only credit} & \textbf{SPRM-Qwen3-4B (ours)} & \checkmark & \checkmark & \checkmark \\
    \bottomrule
    \end{tabular*}
\end{table*}

The baselines in the main experiments differ not only in architecture and
backbone but also in the source of their process supervision.
\Cref{tab:baseline_supervision_compare} separates these differences along three
axes that are central to our setting: whether the process signal is constructed
automatically, whether it avoids external annotators, judges, or curated
intermediate labels, and whether it explicitly assigns step credit by marginal
contribution.
Human-labeled and curated PRMs provide strong process signals but rely on
external process supervision.
Monte Carlo labeling and outcome broadcasting automate part of the labeling
pipeline, but their targets do not isolate each step's marginal contribution to
the final outcome.
First-error boundary models avoid curated intermediate scores, yet they supervise
boundary localization rather than a marginal credit allocation.
SPRM is the only regime in this comparison that derives process labels from
outcome-only trajectories while explicitly estimating marginal contribution.

\subsection{Evaluation Metrics}
\label{sec:appendix_eval_metrics}

We summarize the evaluation metrics used for all benchmarks in the main text
and appendix.
ProcessBench \citep{zheng2025processbench} evaluates first-error localization.
We report \textit{Err}, \textit{Cor}, and F1 to capture first-error
localization, all-correct recognition, and their aggregate trade-off.
PRMBench \citep{DBLP:prmbench} evaluates fine-grained preference behavior
through \textit{Simplicity}, \textit{Soundness}, and \textit{Sensitivity}, which
measure responses to redundant steps, factual or logical errors, and condition
traps or multiple solution paths, respectively.
For Best-of-N inference, we follow process-reward reranking
\citep{lightman2023lets} and report final-answer accuracy across problem
difficulties and search ranges, averaged over 10 random seeds.

\subsection{Computation Cost Comparison}
\label{sec:appendix_compute_cost}

\begin{figure*}[t]
\centering
\begin{subfigure}[t]{0.49\textwidth}
\centering
\includegraphics[width=\linewidth]{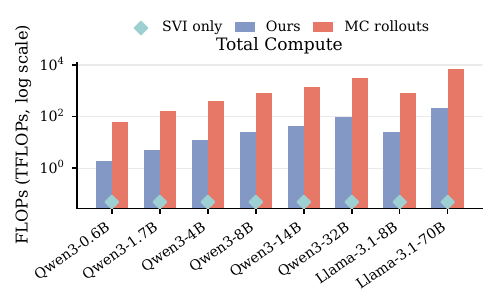}
\caption{Total FLOPs.}
\label{fig:svi_mc_flops_total}
\end{subfigure}
\hfill
\begin{subfigure}[t]{0.49\textwidth}
\centering
\includegraphics[width=\linewidth]{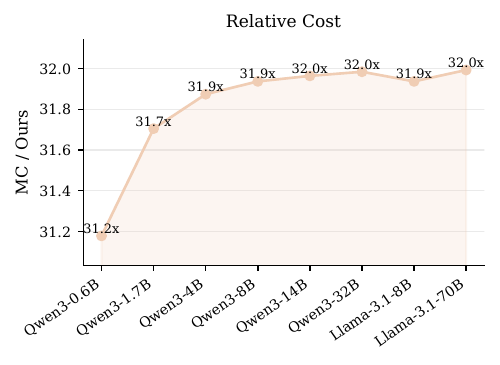}
\caption{Relative MC cost.}
\label{fig:svi_mc_flops_relative}
\end{subfigure}
\caption{Estimated compute for SVI credit assignment versus MC rollout-based
credit estimation.
Blue and red bars show complete total FLOPs for ours and MC, respectively; the
cyan diamond is the SVI-only overhead, not a separate complete method.
MC uses 32 rollouts without shared prompt prefill.}
\label{fig:svi_mc_flops}
\end{figure*}

We include an additional FLOPs comparison between SVI and MC-based step-credit
estimation.
The estimate uses the standard dense-LLM approximation
\(F_{\mathrm{LLM}}=2PT\), where \(P\) is the active parameter count and \(T\)
is the number of processed tokens.
For SVI, the cost is one LLM trajectory pass plus BiLSTM SVI attribution; for
MC, the cost is 32 repeated rollouts of the same prompt-and-solution length.
We use 1024 prompt tokens, 512 solution tokens, 10 reasoning steps, final-layer
step features of dimension 2560, a one-layer BiLSTM with hidden size 512, and
64 SVI integration steps.
\Cref{fig:svi_mc_flops} reports the resulting estimates.
In the total-compute panel, the blue bars and red bars are complete method
costs, while the cyan diamond is only the BiLSTM SVI overhead inside the blue
bar.

\begin{table}[t]
\centering
\footnotesize
\tabtight
\caption{Representative FLOPs breakdown for Qwen3-4B.}
\label{tab:svi_overhead_breakdown}
\begin{tabular}{@{}p{0.44\columnwidth}p{0.22\columnwidth}p{0.25\columnwidth}@{}}
\toprule
\textbf{Quantity} & \textbf{FLOPs} & \textbf{Role in figure} \\
\midrule
Ours total & 12.3 TF & Blue bar \\
MC total & 393 TF & Red bar \\
BiLSTM SVI overhead & 48.6 GF & Cyan diamond \\
One BiLSTM forward & 252 MF & Cost unit \\
\bottomrule
\end{tabular}
\end{table}

As shown in \Cref{tab:svi_overhead_breakdown}, the BiLSTM SVI component is only
\(48.6\) GFLOPs under this setting, so the SVI total is dominated by the single
backbone trajectory pass.
By contrast, MC scales linearly with the rollout budget, making it roughly
32\(\times\) more expensive across the tested backbones.

\subsection{Additional Analysis}
\label{sec:appendix_additional_analysis}

\subsubsection{Value Function Architecture Analysis}

\begin{figure*}[t]
\centering
\includegraphics[width=\textwidth]{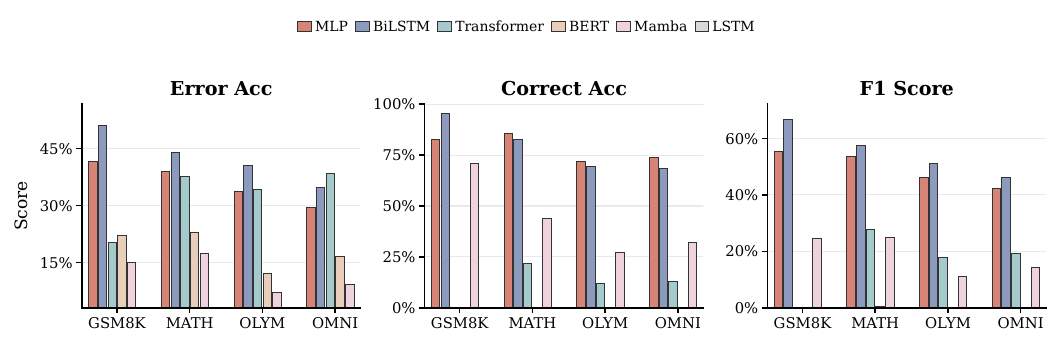}
\caption{Trajectory value function architecture selection across all metrics.
Different value function architectures are compared under a matched optimization setup.}
\label{fig:stage1_model_selection}
\end{figure*}

\Cref{fig:stage1_model_selection} compares lightweight differentiable
trajectory value function architectures by the downstream ProcessBench
performance of the PRM trained from their induced step signals.
For all non-BERT value functions, the input is the encoded sequence
\(X_i=[x_{i,0},x_{i,1},\ldots,x_{i,T_i}]\) used in
\Cref{sec:method_step}, where \(x_{i,0}\) is the
\textit{question\_end} representation and \(x_{i,t}\) is the representation of
step \(s_{i,t}\).
For the BERT baseline, we instead aggregate SVI gradients over each step's
token span at the embedding level.
For the Transformer value function, we test max pooling, mean pooling, and
attention-weighted fusion, and report the attention-weighted variant.
For the MLP value function, we test max and mean pooling and report mean pooling.

The results suggest that compressed step-level representations are necessary:
BERT obtains nearly zero F1, indicating that direct embedding-level attribution
introduces mostly noisy and ineffective gradients.
BiLSTM consistently outperforms LSTM across metrics, showing that bidirectional
context helps model cross-step dependencies for final-reward prediction.
MLP performs closest to BiLSTM, suggesting that feed-forward scoring preserves
substantial information for gradient integration, while BiLSTM provides a better
trade-off between information preservation and task-relevant sequence modeling.

\subsubsection{Step Representation Analysis}
\label{step_features_selection}

\begin{figure*}[t]
    \centering
    \begin{subfigure}[t]{0.32\textwidth}
        \centering
        \includegraphics[width=\linewidth]{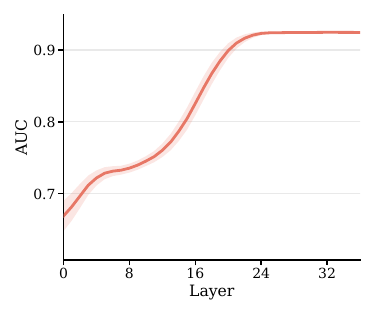}
        \caption{Layer-wise AUC.}
        \label{fig:mlp_layers_auc}
    \end{subfigure}
    \hfill
    \begin{subfigure}[t]{0.32\textwidth}
        \centering
        \includegraphics[width=\linewidth]{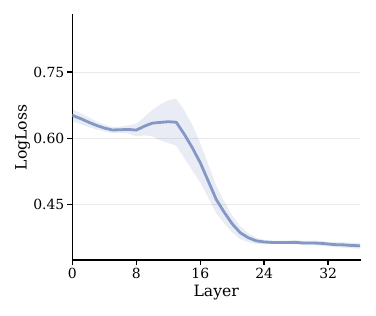}
        \caption{Layer-wise log-loss.} 
        \label{fig:mlp_layers_loss}
    \end{subfigure}
    \hfill
    \begin{subfigure}[t]{0.32\textwidth}
        \centering
        \includegraphics[width=\linewidth]{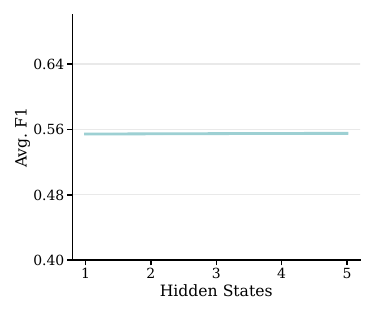}
        \caption{F1 vs. hidden states.} 
        \label{fig:multi_hidden}
    \end{subfigure}

    \caption{Input feature selection: (a) AUC analysis, (b) log-loss analysis,
    and (c) F1 variation on ProcessBench.}
    \label{fig:input_feature_selection}
\end{figure*}

In this section, we investigate three key questions regarding the step
representation \(x_{i,t}\in\mathbb{R}^{d}\): which single hidden layer provides
the optimal representation, how many layers should be combined, and whether
prefix differencing is beneficial.
To determine which layers best capture step semantics, we conduct linear-probe
experiments by training the trajectory value function on hidden states
extracted from various layers.
Furthermore, we train PRMs using varying numbers of hidden layers and evaluate
the impact of differential versus raw inputs through qualitative case studies.

As illustrated in \Cref{fig:input_feature_selection}(a) and (b),
linear-probe experiments reveal that representations from deeper layers
(those closer to the output) are significantly more effective for final-reward
prediction by the trajectory value function, with both AUC and log-loss
converging to optimal levels in the final layers.
Moreover, \Cref{fig:input_feature_selection}(c) demonstrates that
concatenating hidden states from multiple top layers yields no observable
improvement in the downstream PRM F1 score.
Consequently, we use the single final-layer hidden state for step
representation, maximizing computational efficiency without compromising
performance.

To address the differential-input question, we analyze a qualitative reasoning
case where a critical mistake occurs at Step 1.
Although subsequent steps maintain local computational consistency, they stem
from a flawed premise and must ultimately be judged as incorrect.
Our analysis reveals that raw hidden states better preserve global logical
coherence by retaining full contextual history, whereas prefix differencing
aggressively discards crucial cross-step dependencies.

\subsubsection{SVI Baseline Analysis}
\label{svi_baseline_selection}
To choose the SVI baseline, we consider two options: (i) using a first-order
forward-shifted sequence as the baseline, and (ii) using a single semantically
empty vector \(b\) for all encoded positions, as in \Cref{sec:method_step}.

The motivation for option (i) is to capture incremental step contribution.
In this case, for input \(X_i=[x_{i,0},\ldots,x_{i,T_i}]\), the baseline
sequence is
\(B_i^{\mathrm{shift}}=[b,x_{i,0},\ldots,x_{i,T_i-1}]\), where \(b\) is the
semantically empty vector used for the \textit{question\_end} position.
Let \(\delta_{i,t}^{\mathrm{cre,shift}}\) denote the SVI credit computed with
this shifted baseline.
SVI completeness then gives:

\begin{equation}
\label{eq:baseline_selection_delta}
\sum_{t=0}^{T_i}\delta_{i,t}^{\mathrm{cre,shift}}
=
v(X_i) - v(B_i^{\mathrm{shift}}).
\end{equation}

For the trajectory value function, the objective is to predict final reward from
step-level features, so \(v(B_i^{\mathrm{shift}})\) can be interpreted as the
value prediction of a right-shifted trajectory.
The resulting SVI signal therefore attributes steps relative to this shifted
context, which is semantically misaligned with our intended notion of stepwise
marginal contribution.
For this reason, we do not adopt option (i).

For option (ii), summing per-step SVI signals gives:
\begin{equation}
\label{eq:baseline_selection_zero}
\sum_{t=0}^{T_i}\delta_{i,t}^{\mathrm{cre}}
= 
v(X_i) - v(B_i),
\end{equation}
where \(B_i=[b,\ldots,b]\).

According to the SVI definition, each step attribution reflects its contribution
relative to the baseline prediction.
In \citep{sundararajan2017axiomatic}, \(b\) is set to the
all-zero vector for image attribution, with the goal of using a minimally
informative state to avoid biased attributions.
In process-reward modeling with binary outcome signals \((0/1)\), we define a
minimally informative baseline as one producing near-neutral output, \ie,
\(v(B_i)\approx 0.5\), so that attribution signs directly
indicate whether a step is beneficial or harmful.
Comparing an all-zero choice of \(b\) with random-noise choices, we find that
the all-zero baseline yields outputs stably close to \(0.5\), whereas random
baselines are less stable.
We therefore adopt the all-zero vector for \(b\), which preserves completeness
and the semantics of feature-level SVI attribution, and best matches the
Aumann-Shapley perspective.

\subsubsection{CTI Score Analysis}
\label{sec:appendix_cti_score_analysis}

\begin{figure}[!t]
\centering
\captionsetup[subfigure]{justification=centering}
\begin{subfigure}[t]{0.47\textwidth}
    \centering
    \includegraphics[height=0.22\textheight, keepaspectratio]{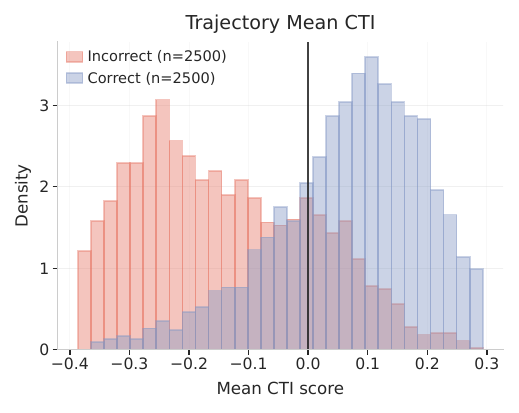}
    \caption{Mean CTI score by reward group.}
    \label{fig:cti_score_hist}
\end{subfigure}
\hfill
\begin{subfigure}[t]{0.47\textwidth}
    \centering
    \includegraphics[height=0.22\textheight, keepaspectratio]{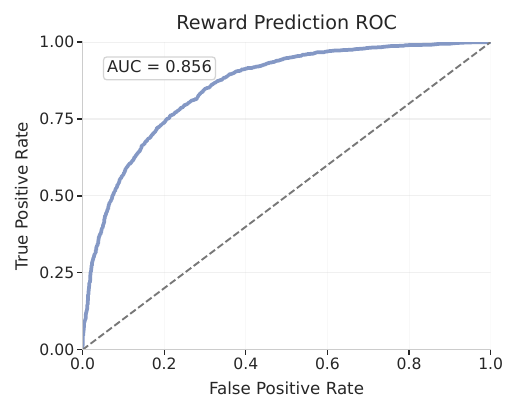}
    \caption{ROC curve using mean CTI score.}
    \label{fig:cti_score_roc}
\end{subfigure}
\vspace{-1mm}
\caption{CTI score analysis with two trajectory-level views.}
\label{fig:cti_score_analysis}
\vspace{-4mm}
\end{figure}

To empirically validate the efficacy of the proposed CTI score
\(\delta_{i,t}^{\mathrm{cti}}\), we analyze its trajectory-level distribution
and predictive capacity.
As defined in \Cref{sec:method_cti}, \(\delta_{i,t}^{\mathrm{cti}}\) acts as a
cross-trajectory inherent score, approximating the relative success tendency of
specific semantic patterns via nearest-neighbor voting.

\Cref{fig:cti_score_analysis}(a) illustrates the distribution of the
mean CTI score, aggregated over all steps within an individual trajectory,
stratified by the final episodic return, with \(n=2500\) for each group.
The empirical distributions exhibit a clear and substantial separation.
Successful trajectories are predominantly characterized by a positive mean
\(\delta^{\mathrm{cti}}\), whereas unsuccessful trajectories consistently
yield negative values.
This observation aligns seamlessly with our theoretical premise: state
representations associated with successful outcomes reliably retrieve
high-reward neighbors from the cross-trajectory candidate pool \(\Omega\),
confirming that \(\delta^{\mathrm{cti}} > 0\) effectively encodes a
higher-than-average propensity for success.

To rigorously quantify the discriminative power of this score,
\Cref{fig:cti_score_analysis}(b) presents the Receiver Operating
Characteristic (ROC) curve for predicting the binary final reward
\(r_i \in \{0, 1\}\) solely using the trajectory-mean \(\delta^{\mathrm{cti}}\).
The estimator achieves an Area Under the Curve (AUC) of 0.856, demonstrating
strong predictive performance significantly above random chance.

Collectively, these results substantiate the utility of our non-parametric
voting estimator.
By surfacing robust, cross-sample statistical constraints, the CTI score
provides a highly discriminative empirical signal that complements the
intra-trajectory step credits \(\delta_{i,t}^{\mathrm{cre}}\).

\subsection{Full Best-of-N Results}
\label{sec:appendix_bon_results}

This section reports the complete AceMath-RewardBench Best-of-N evaluation
omitted from the main text for space.
The full comparison includes dataset-wise PRM reranking results across all
released reasoning-trajectory sets and candidate-pool sizes under the same
sampling protocol.
For SPRM, each entry reports the best accuracy over \texttt{soft}/\texttt{hard}
score transformations and \texttt{min}/\texttt{mean} step aggregation.
For PRM baselines, each entry reports the best accuracy over the same
sorting variants when multiple variants are available.
Pass@\(K\) and majority vote are reported as reference selection baselines
above the PRM reranking models.
We omit the \texttt{@1} column because all methods share the same greedy
accuracy at \(N{=}1\).
In each remaining column, boldface marks the best score among reranking
baselines and PRMs (\ie{}, excluding the Pass@\(K\) upper bound).

\begin{table}[H]
\centering
\footnotesize
\setlength{\tabcolsep}{3.4pt}
\renewcommand{\arraystretch}{0.88}
\caption{Complete Best-of-N accuracy on CollegeMath, Gaokao2023en, GSM8K, and
MATH500.}
\label{tab:appendix_bon_collegemath}
\label{tab:appendix_bon_gaokao2023en}
\label{tab:appendix_bon_gsm8k}
\label{tab:appendix_bon_math500}
\begin{tabular*}{\textwidth}{@{\extracolsep{\fill}}lrrrrrr@{}}
\toprule
\textbf{Model} & \textbf{@2} & \textbf{@4} & \textbf{@8} & \textbf{@16} & \textbf{@32} & \textbf{@64} \\
\midrule\midrule
\multicolumn{7}{@{}l}{\textbf{CollegeMath}} \\
\midrule
Pass@K (upper bound) & 70.9 & 77.9 & 82.7 & 86.1 & 87.4 & 89.0 \\
Majority vote & 58.8 & 67.2 & 70.1 & 71.3 & 71.9 & \textbf{72.0} \\
\midrule
Llama3.1-PRM-D & 59.6 & 60.0 & 60.1 & 59.4 & 57.9 & 55.9 \\
Llama3.1-PRM-M & 59.1 & 59.2 & 59.2 & 57.3 & 55.9 & 53.2 \\
Math-Shepherd & 60.5 & 60.8 & 60.8 & 60.2 & 60.6 & 57.0 \\
Qwen2.5-Math-7B-PRM800K & 59.1 & 59.2 & 59.2 & 59.2 & 59.2 & 59.2 \\
Skywork-o1-PRM & 61.0 & 63.1 & 63.9 & 64.6 & 63.1 & 61.0 \\
ReasonEval-7B & 58.5 & 58.6 & 58.6 & 58.6 & 56.9 & 54.5 \\
PQM & 59.2 & 59.6 & 59.8 & 59.6 & 58.9 & 57.8 \\
OVM & 58.6 & 58.8 & 58.6 & 58.2 & 57.1 & 56.0 \\
\textbf{SPRM} & \textbf{66.1} & \textbf{69.4} & \textbf{70.5} & \textbf{71.5} & \textbf{72.0} & \textbf{72.0} \\
\midrule\midrule
\multicolumn{7}{@{}l}{\textbf{Gaokao2023en}} \\
\midrule
Pass@K (upper bound) & 62.7 & 74.6 & 79.9 & 82.7 & 85.0 & 86.0 \\
Majority vote & 45.1 & 59.1 & 65.9 & 70.7 & 72.8 & 72.0 \\
\midrule
Llama3.1-PRM-D & 45.0 & 44.8 & 44.9 & 44.9 & 44.9 & 44.9 \\
Llama3.1-PRM-M & 46.6 & 46.6 & 46.6 & 46.6 & 46.6 & 45.0 \\
Math-Shepherd & 48.9 & 49.6 & 49.6 & 49.6 & 49.6 & 49.6 \\
Qwen2.5-Math-7B-PRM800K & 46.3 & 46.5 & 46.5 & 46.5 & 46.5 & 46.5 \\
Skywork-o1-PRM & 52.9 & 57.8 & 59.8 & 61.6 & 63.0 & 64.1 \\
ReasonEval-7B & 45.9 & 46.0 & 46.0 & 46.0 & 46.0 & 46.0 \\
PQM & 47.8 & 49.5 & 50.3 & 50.6 & 49.8 & 49.0 \\
OVM & 47.0 & 47.8 & 48.2 & 47.6 & 47.1 & 46.8 \\
\textbf{SPRM} & \textbf{59.1} & \textbf{68.4} & \textbf{72.1} & \textbf{73.0} & \textbf{73.2} & \textbf{73.0} \\
\midrule\midrule
\multicolumn{7}{@{}l}{\textbf{GSM8K}} \\
\midrule
Pass@K (upper bound) & 94.7 & 98.1 & 99.2 & 100.0 & 100.0 & 100.0 \\
Majority vote & 87.1 & 94.2 & 96.2 & 96.9 & \textbf{97.3} & \textbf{98.3} \\
\midrule
Llama3.1-PRM-D & 87.0 & 87.0 & 87.0 & 87.0 & 87.0 & 87.0 \\
Llama3.1-PRM-M & 87.9 & 88.1 & 88.1 & 88.1 & 88.1 & 88.1 \\
Math-Shepherd & 88.1 & 88.5 & 88.5 & 88.5 & 88.5 & 88.5 \\
Qwen2.5-Math-7B-PRM800K & 87.2 & 87.4 & 87.4 & 87.4 & 87.4 & 87.4 \\
Skywork-o1-PRM & 90.1 & 92.4 & 93.0 & 93.7 & 94.3 & 94.2 \\
ReasonEval-7B & 87.1 & 87.1 & 87.1 & 87.1 & 87.1 & 87.1 \\
PQM & 87.6 & 88.3 & 88.6 & 88.7 & 88.4 & 88.0 \\
OVM & 87.2 & 87.4 & 87.5 & 87.2 & 86.9 & 86.5 \\
\textbf{SPRM} & \textbf{92.3} & \textbf{95.2} & \textbf{96.8} & \textbf{97.2} & 97.2 & 97.0 \\
\midrule\midrule
\multicolumn{7}{@{}l}{\textbf{MATH500}} \\
\midrule
Pass@K (upper bound) & 81.5 & 91.1 & 93.9 & 95.6 & 96.4 & 97.0 \\
Majority vote & 65.8 & 79.4 & 86.2 & 88.5 & \textbf{89.7} & \textbf{90.0} \\
\midrule
Llama3.1-PRM-D & 66.4 & 67.6 & 67.7 & 67.8 & 67.8 & 67.8 \\
Llama3.1-PRM-M & 66.4 & 66.8 & 66.8 & 66.8 & 66.8 & 66.8 \\
Math-Shepherd & 68.3 & 68.8 & 68.8 & 68.8 & 68.8 & 68.8 \\
Qwen2.5-Math-7B-PRM800K & 67.5 & 67.8 & 67.8 & 67.8 & 67.8 & 67.8 \\
Skywork-o1-PRM & 73.6 & 78.4 & 79.7 & 80.8 & 81.1 & 81.3 \\
ReasonEval-7B & 66.4 & 66.7 & 66.7 & 66.7 & 66.7 & 65.9 \\
PQM & 67.1 & 68.4 & 69.1 & 69.4 & 68.5 & 67.4 \\
OVM & 66.4 & 66.9 & 67.2 & 66.9 & 66.4 & 65.9 \\
\textbf{SPRM} & \textbf{77.8} & \textbf{86.4} & \textbf{88.8} & \textbf{88.8} & 89.0 & \textbf{90.0} \\
\bottomrule
\end{tabular*}
\end{table}

\begin{table}[H]
\centering
\footnotesize
\setlength{\tabcolsep}{3.4pt}
\renewcommand{\arraystretch}{0.88}
\caption{Complete Best-of-N accuracy on Minerva Math, MMLU-STEM, and
OlympiadBench.}
\label{tab:appendix_bon_minervamath}
\label{tab:appendix_bon_mmlustem}
\label{tab:appendix_bon_olympiadbench}
\begin{tabular*}{\textwidth}{@{\extracolsep{\fill}}lrrrrrr@{}}
\toprule
\textbf{Model} & \textbf{@2} & \textbf{@4} & \textbf{@8} & \textbf{@16} & \textbf{@32} & \textbf{@64} \\
\midrule\midrule
\multicolumn{7}{@{}l}{\textbf{Minerva Math}} \\
\midrule
Pass@K (upper bound) & 41.4 & 53.8 & 62.8 & 68.9 & 72.1 & 75.0 \\
Majority vote & 28.6 & 38.2 & 44.5 & \textbf{48.7} & \textbf{51.2} & \textbf{52.4} \\
\midrule
Llama3.1-PRM-D & 30.1 & 30.3 & 30.3 & 30.3 & 30.3 & 30.3 \\
Llama3.1-PRM-M & 29.3 & 29.5 & 29.5 & 29.5 & 29.5 & 29.5 \\
Math-Shepherd & 30.1 & 30.4 & 30.4 & 30.4 & 30.4 & 30.4 \\
Qwen2.5-Math-7B-PRM800K & 29.3 & 29.4 & 29.4 & 29.4 & 29.4 & 29.4 \\
Skywork-o1-PRM & 30.5 & 32.3 & 33.9 & 35.4 & 36.5 & 37.6 \\
ReasonEval-7B & 28.9 & 29.1 & 29.1 & 29.1 & 29.1 & 29.1 \\
PQM & 28.9 & 29.5 & 29.8 & 29.6 & 29.1 & 28.6 \\
OVM & 28.6 & 28.8 & 28.9 & 28.6 & 28.4 & 28.2 \\
\textbf{SPRM} & \textbf{37.0} & \textbf{42.4} & \textbf{45.6} & 45.3 & 46.2 & 46.0 \\
\midrule\midrule
\multicolumn{7}{@{}l}{\textbf{MMLU-STEM}} \\
\midrule
Pass@K (upper bound) & 76.4 & 90.9 & 97.3 & 99.4 & 99.8 & 100.0 \\
Majority vote & 53.8 & 69.4 & 76.2 & 81.1 & 83.9 & 85.8 \\
\midrule
Llama3.1-PRM-D & 54.4 & 54.5 & 54.5 & 54.5 & 54.5 & 54.5 \\
Llama3.1-PRM-M & 54.4 & 54.6 & 54.6 & 54.6 & 54.6 & 54.6 \\
Math-Shepherd & 55.8 & 56.1 & 56.1 & 56.1 & 56.1 & 53.6 \\
Qwen2.5-Math-7B-PRM800K & 54.4 & 54.4 & 54.4 & 54.4 & 54.4 & 54.4 \\
Skywork-o1-PRM & 54.6 & 54.9 & 55.8 & 55.9 & 55.8 & 55.8 \\
ReasonEval-7B & 53.8 & 53.7 & 53.7 & 53.6 & 53.4 & 53.4 \\
PQM & 54.2 & 54.7 & 55.0 & 54.8 & 54.3 & 53.9 \\
OVM & 53.9 & 54.1 & 54.2 & 53.9 & 53.6 & 53.3 \\
\textbf{SPRM} & \textbf{67.4} & \textbf{76.2} & \textbf{80.6} & \textbf{83.6} & \textbf{86.4} & \textbf{88.0} \\
\midrule\midrule
\multicolumn{7}{@{}l}{\textbf{OlympiadBench}} \\
\midrule
Pass@K (upper bound) & 19.2 & 29.1 & 37.2 & 44.9 & 50.7 & 55.0 \\
Majority vote & 12.2 & 16.4 & 19.8 & 22.2 & 22.6 & 23.6 \\
\midrule
Llama3.1-PRM-D & 11.1 & 11.4 & 11.0 & 9.6 & 8.7 & 7.6 \\
Llama3.1-PRM-M & 11.4 & 12.7 & 12.3 & 9.8 & 9.2 & 8.0 \\
Math-Shepherd & 13.8 & 14.6 & 14.6 & 14.6 & 14.6 & 14.6 \\
Qwen2.5-Math-7B-PRM800K & 12.2 & 12.2 & 12.2 & 12.2 & 12.2 & 12.2 \\
Skywork-o1-PRM & 12.4 & 12.7 & 13.2 & 14.3 & 15.0 & 17.3 \\
ReasonEval-7B & 12.0 & 12.3 & 11.8 & 12.1 & 11.8 & 9.1 \\
PQM & 12.4 & 12.8 & 13.0 & 12.8 & 12.4 & 11.8 \\
OVM & 12.2 & 12.4 & 12.5 & 12.3 & 12.2 & 11.7 \\
\textbf{SPRM} & \textbf{16.1} & \textbf{19.6} & \textbf{20.5} & \textbf{23.1} & \textbf{24.4} & \textbf{26.0} \\
\bottomrule
\end{tabular*}
\end{table}

\subsection{Same-Backbone Supervision Comparison}
\label{sec:appendix_same_backbone}

To isolate the effect of supervision, we retrain OVM, Monte Carlo,
LLM-judge, and SPRM variants with the same Qwen3-4B-Instruct-2507 backbone
on a shared, fixed subset of 20K Math-Shepherd trajectories.
The remaining architecture, optimization, preprocessing, and ProcessBench
evaluation settings are kept identical to the original implementation.
The human-label reference uses the same backbone but a randomly sampled
20K subset of PRM800K, the available human process-labeled source,
and therefore does not control for training data.
Unlike the broader comparison in \Cref{tab:processbench_res}, which includes
publicly available PRMs with different backbones and supervision regimes,
this experiment controls the backbone and, except for the human reference,
the training subset.

OVM broadcasts terminal outcome labels to all steps, whereas Monte Carlo
uses rollout-estimated step labels.
The LLM-judge baseline uses binary labels from DeepSeek-V3.2, which receives
the problem, previous steps, and current step and answers only
\texttt{Yes} or \texttt{No}.
SPRM constructs marginal-contribution supervision from terminal outcome
rewards without human step labels, external step judges, or rollout-derived
step labels.

\begin{table}[htbp]
\centering
\footnotesize
\tabtight
\caption{Same-backbone supervision comparison on ProcessBench (F1).
All variants use Qwen3-4B-Instruct-2507 and 20K training trajectories.
Math-Shepherd rows share the same subset; the human-label reference uses
PRM800K instead. Boldface marks the best score in each column.}
\label{tab:same_backbone}
\begin{tabular*}{\textwidth}{@{\extracolsep{\fill}}llrrrrr@{}}
\toprule
\textbf{Supervision} & \textbf{Training source}
    & \textbf{GSM8K} & \textbf{MATH} & \textbf{OlympiadBench}
    & \textbf{OmniMath} & \textbf{Avg. F1} \\
\midrule
Outcome broadcast (OVM) & Math-Shepherd
    & 8.6 & 46.5 & 32.9 & 30.1 & 29.5 \\
Monte Carlo labels & Math-Shepherd
    & 63.2 & 37.4 & 21.2 & 17.8 & 34.9 \\
Human labels & PRM800K
    & 66.6 & 55.2 & 38.4 & 31.4 & 47.9 \\
LLM-judge labels & Math-Shepherd
    & 68.2 & 58.5 & 51.0 & 50.4 & 57.0 \\
\midrule
\textbf{SPRM (ours)} & Math-Shepherd
    & \textbf{68.6} & \textbf{59.2} & \textbf{53.6}
    & \textbf{52.6} & \textbf{58.5} \\
\bottomrule
\end{tabular*}
\end{table}

As shown in \Cref{tab:same_backbone}, SPRM achieves the highest F1 on all four
ProcessBench splits, with an average F1 of 58.5 versus 57.0 for the strongest
baseline, LLM-judge supervision.
It exceeds outcome broadcasting and Monte Carlo supervision by 29.0 and
23.6 average-F1 points, respectively, using only 20K training trajectories.
These results support the benefit of marginal-contribution supervision
under a shared backbone and training subset.

\subsection{Downstream Reinforcement Learning}
\label{sec:appendix_downstream_rl}

We evaluate whether SPRM improves downstream policy optimization by training
Qwen3-1.7B-Base on MATH and evaluating on MATH-500.
We compare RLOO with terminal correctness rewards, RLOO + OVM with an
outcome-broadcast process reward model, and RLOO + SPRM with our
marginal-contribution process reward model.
All three settings share the initial policy and training setup.
OVM and SPRM use the same reward-model backbone and the same min-form
reward integration with RLOO, allowing a controlled comparison of their
process supervision signals.
We use the default min-form hyperparameters for both process-reward variants.

We optimize the policy with AdamW using a constant learning rate of
\(5\times10^{-6}\), weight decay 0.01, and LoRA rank 64 with alpha 128.
Each training batch contains 32 prompts with eight rollouts per prompt.
We use a policy-update mini-batch size of 32, one policy-optimization epoch
per rollout batch, and a PPO clipping ratio of 0.2.
Training rollouts use temperature 1.2 and top-\(p\) 1.0, with prompt and
response lengths each capped at 1,024 tokens.

We repeat RL training with three independent random seeds for each method.
For each run, we evaluate mean@8 accuracy, averaging correctness over eight
sampled responses per problem.
\Cref{tab:downstream_rl} reports the mean and standard deviation across
independent RL training seeds at each reported training step.
The shared initial policy achieves 55.0\% mean@8 accuracy before RL training.


\begin{table}[htbp]
\centering
\footnotesize
\tabtight
\caption{Downstream RL results on MATH-500 with Qwen3-1.7B-Base trained
on MATH. Entries report mean@8 accuracy (\%), as mean $\pm$ standard
deviation across three independent RL training seeds at each training step.
OVM and SPRM use the same min-form reward integration.
Boldface marks the highest mean in each row.}
\label{tab:downstream_rl}

\begin{tabular*}{\linewidth}{@{\extracolsep{\fill}}rccc@{}}
\toprule
\textbf{Training step}
    & \textbf{RLOO}
    & \textbf{RLOO + OVM}
    & \textbf{RLOO + SPRM} \\
\midrule
25
    & 59.4 $\pm$ 1.7
    & 54.2 $\pm$ 1.1
    & \textbf{60.0} $\pm$ 1.5 \\
50
    & 63.0 $\pm$ 0.5
    & 57.2 $\pm$ 0.8
    & \textbf{63.4} $\pm$ 1.1 \\
75
    & 63.6 $\pm$ 0.6
    & 57.8 $\pm$ 1.2
    & \textbf{66.2} $\pm$ 0.7 \\
100
    & 64.4 $\pm$ 0.8
    & 57.4 $\pm$ 0.8
    & \textbf{68.0} $\pm$ 0.9 \\
125
    & 65.6 $\pm$ 0.7
    & 56.4 $\pm$ 1.9
    & \textbf{69.2} $\pm$ 0.7 \\
150
    & 65.2 $\pm$ 1.2
    & 58.2 $\pm$ 1.3
    & \textbf{70.0} $\pm$ 0.8 \\
\bottomrule
\end{tabular*}
\end{table}

At training step 150, RLOO + SPRM reaches 70.0\% mean@8 accuracy,
exceeding RLOO by 4.8 percentage points and RLOO + OVM by 11.8 points.
RLOO + SPRM has the highest mean accuracy at every reported training step.
Its advantage over OVM under the same reward integration supports the
usefulness of marginal-contribution supervision for downstream policy
optimization in this setting.

\clearpage

\section{Implementation and Reproducibility Details}
\label{sec:appendix_impl_details}

This appendix section collects implementation and runtime details for
reproducing the reported runs.
The empirical analyses and additional result tables are kept separately in
\Cref{sec:appendix_additional_experiments}.

\subsection{Component Implementation}

Our implementation consists of three components: trajectory value function
training, step-level attribution construction, and final PRM training.
Unless otherwise stated, all components use Qwen3-4B-Instruct-2507 as the
backbone.

\paragraph{Trajectory value function training}
We randomly sample 220K trajectories from Math-Shepherd and extract final-layer
hidden states for the question boundary and reasoning steps as encoded
positions.
The question-boundary representation is kept as the zeroth encoded position, so
each trajectory is represented as
\([x_{i,0},x_{i,1},\ldots,x_{i,T_i}]\), where \(x_{i,0}\) is
\textit{question\_end}.
Each representation has dimension \(d=2560\), matching the final hidden size of
Qwen3-4B-Instruct-2507.
The trajectory value function is a single-layer bidirectional LSTM with hidden
size 512 in each direction, followed by a linear prediction head that outputs
one trajectory logit.
It is trained with \(\mathrm{BCEWithLogitsLoss}\) on the binary final reward,
using post-padding and packed sequences so that padded positions are ignored.
We train with batch size 128 and learning rate \(1\times10^{-4}\), with early
stopping using patience 5, and split the data into train/validation/test sets
with ratio \(0.8/0.1/0.1\).

\paragraph{Step-level attribution construction}
We compute SVI step attribution with an all-zero baseline and 64 integration
steps, using the trajectory value function probability output as the attribution
target.
For each trajectory, the encoded sequence is treated as a single differentiable
input, and each step credit is obtained by summing the SVI attribution over the
hidden dimensions of the corresponding step representation.
To construct the step-credit supervision mask, we compare the sign of this SVI
credit with the sign of the model-counterfactual consistency score
\(R_{i,t}^{\mathrm{count}}\).
For the local interaction window in \Cref{eq:pairwise_counterfactual}, we use
\(w=3\) and \(\beta=0.5\).
The label for the step-credit head is used only when the two signs agree; the
gate threshold is 0, so no additional magnitude filtering is applied.
The cross-trajectory inherent score is built through semantic nearest-neighbor
retrieval across trajectories.
We L2-normalize all step representations and retrieve neighbors by cosine
similarity, excluding steps from the same trajectory to avoid outcome leakage.
For each step, we search over 400 candidate neighbors, keep the top 50 valid
neighbors, and compute similarity-weighted reward statistics using ReLU-based
weights.
The CTI score is the resulting neighborhood success rate minus the dataset base
success rate.

\paragraph{PRM training}
The constructed step-credit and CTI signals are then used to supervise the final
PRM.
We insert a special \texttt{<ST\_END>} token after the question boundary and
after every reasoning step, and train both value heads only at these boundary
positions.
The first head predicts the binarized SVI step-credit label under the
\(R^{\mathrm{count}}\)-consistency mask, while the second head predicts the
binarized CTI label whenever the CTI score is finite.
Both labels use threshold 0, assigning positive scores to label 1 and nonpositive
scores to label 0.
The total objective is the sum of the two masked BCE losses with equal weights.
We perform LoRA fine-tuning on the backbone with target modules
\texttt{q\_proj}, \texttt{k\_proj}, \texttt{v\_proj}, \texttt{o\_proj},
\texttt{gate\_proj}, \texttt{up\_proj}, and \texttt{down\_proj}.
Training uses maximum context length 4096, runs for 1 epoch with learning rate
\(5\times10^{-5}\), and uses global batch size 32.

\subsection{Run Details and Environment}
\label{sec:appendix_run_environment}

This subsection records the runtime environment for the reported reproducible
runs.
Hyperparameters and component-specific training choices are reported above.

\paragraph{Execution environment}
The reported runs use Linux \texttt{5.15.0-164-generic}, Python
\texttt{3.11.14}, CUDA \texttt{12.8} as used by PyTorch, PyTorch
\texttt{2.9.1}, Transformers \texttt{4.56.1}, Accelerate \texttt{1.6.0},
DeepSpeed \texttt{0.18.3}, PEFT \texttt{0.18.0}, and TRL \texttt{0.11.4}.
The dependency constraints require Python \(\geq 3.11,<3.12\).

\paragraph{Hardware and launch}
Experiments are run on two NVIDIA L20 GPUs with 45 GB memory each.
Trajectory value function training, step-level attribution construction, and
final PRM training use
\texttt{torchrun --nproc\_per\_node 2 --master\_port \$MAIN\_PORT}, where
\texttt{MAIN\_PORT} is selected automatically from 29500--29649 to avoid port
conflicts.

\paragraph{Logging and deterministic settings}
Weights \& Biases logging is enabled by default unless
\texttt{WANDB\_DISABLED} is set.
Trajectory value function training and final PRM training use random seed 42;
step-level attribution construction uses
\(\texttt{STAGE2\_DG\_SAMPLE\_SEED}=1234\) for CTI trajectory sampling.
We fix and report these seeds, but do not claim bitwise deterministic
reproduction because multi-GPU execution, cuDNN, DeepSpeed, and fused AdamW may
still introduce small non-determinism.

\clearpage

\newpage
\section*{NeurIPS Paper Checklist}

\begin{enumerate}

\item {\bf Claims}
    \item[] Question: Do the main claims made in the abstract and introduction accurately reflect the paper's contributions and scope?
    \item[] Answer: \answerYes{}
    \item[] Justification: The abstract and introduction state the core claim
    of deriving step-level supervision from outcome-only trajectories,
    and the conclusion keeps the same scope in light of the reported
    ProcessBench, PRMBench, and Best-of-N results.

\item {\bf Limitations}
    \item[] Question: Does the paper discuss the limitations of the work performed by the authors?
    \item[] Answer: \answerYes{}
    \item[] Justification: The conclusion explicitly states that the current
    method assumes verifiable outcome rewards and identifies reward modeling
    for tasks whose answers cannot be reliably verified as an important
    direction for future work.

\item {\bf Theory assumptions and proofs}
    \item[] Question: For each theoretical result, does the paper provide the full set of assumptions and a complete (and correct) proof?
    \item[] Answer: \answerYes{}
    \item[] Justification: The main text states the two lemmas and
    cross-references the appendix, which defines the notation, lists
    assumptions A1--A4, and provides the corresponding proofs and axiomatic
    comparison.

    \item {\bf Experimental result reproducibility}
    \item[] Question: Does the paper fully disclose all the information needed to reproduce the main experimental results of the paper to the extent that it affects the main claims and/or conclusions of the paper (regardless of whether the code and data are provided or not)?
    \item[] Answer: \answerYes{}
    \item[] Justification: The experiments section describes datasets,
    baselines, metrics, and evaluation protocol, while Appendix C provides
    implementation details for value-function training, attribution
    construction, retrieval, PRM training, runtime environment, distributed
    launch, hardware, logging, and seeds.

\item {\bf Open access to data and code}
    \item[] Question: Does the paper provide open access to the data and code, with sufficient instructions to faithfully reproduce the main experimental results, as described in supplemental material?
    \item[] Answer: \answerYes{}
    \item[] Justification: We provide a public code repository at
    \url{https://github.com/by0214/sprm-open},
    while the paper and appendix describe the datasets, baselines, metrics,
    evaluation protocol, implementation details, runtime environment, hardware,
    logging, and seeds needed to reproduce the main experimental results.

\item {\bf Experimental setting/details}
    \item[] Question: Does the paper specify all the training and test details (e.g., data splits, hyperparameters, how they were chosen, type of optimizer) necessary to understand the results?
    \item[] Answer: \answerYes{}
    \item[] Justification: The experiments section and Appendix C specify the
    benchmarks, training data, backbone, evaluation protocol, data split,
    SVI settings, retrieval configuration, PRM training hyperparameters,
    runtime environment, and distributed launch details.

\item {\bf Experiment statistical significance}
    \item[] Question: Does the paper report error bars suitably and correctly defined or other appropriate information about the statistical significance of the experiments?
    \item[] Answer: \answerNo{}
    \item[] Justification: The main benchmark tables do not report error bars
    or formal significance tests, although Best-of-N results are averaged over
    fixed random-seed reranking runs.

\item {\bf Experiments compute resources}
    \item[] Question: For each experiment, does the paper provide sufficient information on the computer resources (type of compute workers, memory, time of execution) needed to reproduce the experiments?
    \item[] Answer: \answerYes{}
    \item[] Justification: The appendix reports the two-GPU L20 setup,
    distributed launch details, runtime environment, and FLOPs comparison
    between SVI and MC rollout-based credit estimation.
    
\item {\bf Code of ethics}
    \item[] Question: Does the research conducted in the paper conform, in every respect, with the NeurIPS Code of Ethics \url{https://neurips.cc/public/EthicsGuidelines}?
    \item[] Answer: \answerYes{}
    \item[] Justification: The work is a methodological study on process
    reward modeling using existing benchmark trajectories and does not involve
    deception, human-subject experimentation, or release practices that
    intentionally conflict with the NeurIPS Code of Ethics.

\item {\bf Broader impacts}
    \item[] Question: Does the paper discuss both potential positive societal impacts and negative societal impacts of the work performed?
    \item[] Answer: \answerYes{}
    \item[] Justification: The conclusion discusses both positive impacts,
    such as reducing dependence on costly external judges, and negative
    impacts, such as strengthening automated reasoning systems in settings
    where persuasive errors can be harmful.
    
\item {\bf Safeguards}
    \item[] Question: Does the paper describe safeguards that have been put in place for responsible release of data or models that have a high risk for misuse (e.g., pre-trained language models, image generators, or scraped datasets)?
    \item[] Answer: \answerNA{}
    \item[] Justification: The paper does not release a new high-risk
    generative model or scraped dataset as part of the current submission.

\item {\bf Licenses for existing assets}
    \item[] Question: Are the creators or original owners of assets (e.g., code, data, models), used in the paper, properly credited and are the license and terms of use explicitly mentioned and properly respected?
    \item[] Answer: \answerYes{}
    \item[] Justification: The paper credits the benchmark datasets, models,
    and prior systems used in the work through the corresponding citations.

\item {\bf New assets}
    \item[] Question: Are new assets introduced in the paper well documented and is the documentation provided alongside the assets?
    \item[] Answer: \answerNA{}
    \item[] Justification: The current submission does not release a new
    dataset, code package, or model artifact as a documented public asset.

\item {\bf Crowdsourcing and research with human subjects}
    \item[] Question: For crowdsourcing experiments and research with human subjects, does the paper include the full text of instructions given to participants and screenshots, if applicable, as well as details about compensation (if any)? 
    \item[] Answer: \answerNA{}
    \item[] Justification: The paper does not involve crowdsourcing or
    research with human subjects.

\item {\bf Institutional review board (IRB) approvals or equivalent for research with human subjects}
    \item[] Question: Does the paper describe potential risks incurred by study participants, whether such risks were disclosed to the subjects, and whether Institutional Review Board (IRB) approvals (or an equivalent approval/review based on the requirements of your country or institution) were obtained?
    \item[] Answer: \answerNA{}
    \item[] Justification: The paper does not involve human-subject research
    requiring IRB review.

\item {\bf Declaration of LLM usage}
    \item[] Question: Does the paper describe the usage of LLMs if it is an important, original, or non-standard component of the core methods in this research? Note that if the LLM is used only for writing, editing, or formatting purposes and does \emph{not} impact the core methodology, scientific rigor, or originality of the research, declaration is not required.
    \item[] Answer: \answerNA{}
    \item[] Justification: LLMs are used only as existing experimental
    backbones or representation sources and are not an important, original,
    or non-standard component of the proposed core methodology.

\end{enumerate}

\end{document}